\documentclass[10pt,twocolumn,letterpaper]{article}

\usepackage{iccv}
\usepackage{times}
\usepackage{epsfig}
\usepackage{graphicx}
\usepackage{amsmath}
\usepackage{amssymb}
\usepackage{enumitem}
\usepackage[pagebackref=true,breaklinks=true,letterpaper=true,colorlinks,bookmarks=false]{hyperref}
\usepackage{xcolor}
\usepackage{subfig}
\usepackage{booktabs}
\usepackage{multirow}
\let\svthefootnote\thefootnote
\newcommand\blankfootnote[1]{
  \let\thefootnote\relax\footnotetext{#1}
  \let\thefootnote\svthefootnote
}
\usepackage[accsupp]{axessibility}
\usepackage{color, colortbl}
\definecolor{Gray}{gray}{0.9}

\newcommand{\sigrob}[1]{\colorbox{blue!15}{#1}}

\iccvfinalcopy

\ificcvfinal\fi

\begin{document}

\title{Benchmarking Low-Shot Robustness to Natural Distribution Shifts}
\author{
$\text{Aaditya Singh}^{\ast,\star,\dagger}$
\qquad
$\text{Kartik Sarangmath}^{\ast}$
\qquad
Prithvijit Chattopadhyay
\qquad
Judy Hoffman \\
{Georgia Institute of Technology}\\
{\tt\small \{singaadi, kartiksarangmath, prithvijit3, judy\}@gatech.edu}\\
}
\maketitle

\begin{abstract}
Robustness to natural distribution shifts has seen remarkable progress thanks to recent pre-training strategies combined with better fine-tuning methods. However, such fine-tuning assumes access to large amounts of labelled data, and the extent to which the observations hold when the amount of training data is not as high remains unknown. We address this gap by performing the first in-depth study of robustness to various natural distribution shifts in different low-shot regimes: spanning datasets, architectures, pre-trained initializations, and state-of-the-art robustness interventions.
Most importantly, we find that there is no single model of choice that is often more robust than others, and existing interventions can fail to improve robustness on some datasets even if they do so in the full-shot regime. We hope that our work will motivate the community to focus on this problem of practical importance.
Our code and low-shot subsets are publicly available at 
\href{https://github.com/Aaditya-Singh/Low-Shot-Robustness/}{this url}.
\end{abstract}

\section{Introduction}
\label{sec:intro}
\blankfootnote{
\textsuperscript{$\ast$}Equal contribution; \textsuperscript{$\star$}Project lead; \textsuperscript{$\dagger$}Currently affiliated with AWS AI Labs, work done prior to joining.
}

In the past decade, Computer Vision has made significant progress due to advanced architectures like Convolutional Neural Networks (CNNs) and Vision Transformers (ViTs), large datasets, and sophisticated training strategies~\cite{he2016deep, kolesnikov2020big, dosovitskiy2020image, radford2021learning}. However, early learning techniques heavily focused their evaluation on ImageNet~\cite{deng2009imagenet} performance, which raised concerns about their ability to generalize to distribution shifts~\cite{geirhos2020shortcut, recht2019imagenet}. To address this, researchers have proposed a wide-range of evaluation datasets~\cite{barbu2019objectnet, wang2019learning, hendrycks2021natural, hendrycks2021many, koh2021wilds} that can be used to measure out-of-distribution (OOD) performance of models trained and validated with in-domain (ID) data.

\begin{figure}[t]
\centering
\includegraphics[width=\columnwidth]{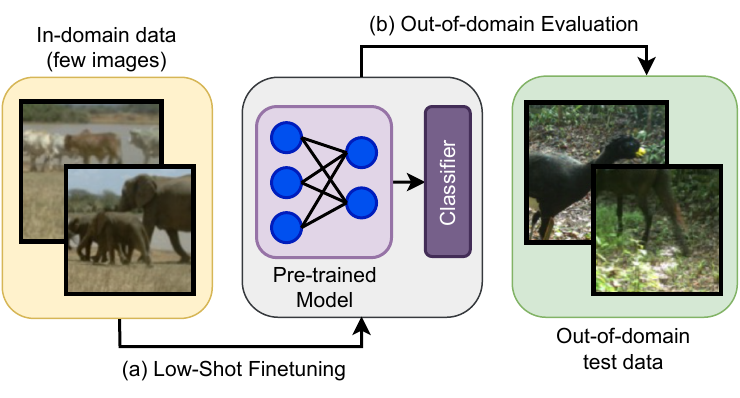}
\caption{\small\textbf{Low-Shot Robustness Setting.} (a)
We assume access to a pre-trained model trained on large-scale datasets such as ImageNet \cite{deng2009imagenet} and limited in-domain images (in the order of thousands) for training. We use different kinds of fine-tuning methods that have been shown to improve robustness when there is typically order of magnitudes higher training data.
(b) We then evaluate the (low-shot) fine-tuned model on out-of-domain (OOD) data.
}
\label{fig:teaser}
\vspace{-15pt}
\end{figure}

Recent methods \cite{he2022masked, assran2022masked, kumar2022fine, wortsman2022robust, wortsman2022model} use self-supervised or large-scale vision-language pre-trained models (such as CLIP~\cite{radford2021learning}) and fine-tune them on fully labelled ID data to achieve impressive performance on such datasets. 
Unfortunately, fine-tuning requires large amounts of data and compute that may not be accessible to most practitioners. Moreover, it can be difficult and expensive to collect and consistently annotate such data, especially in settings like camera traps where images can vary significantly in quality, lighting, and pose (e.g~iWildCam~\cite{beery2020iwildcam}). Such challenges are also echoed by prior work \cite{wiles2022a} and compounded by the fact that many images may belong to rare or endangered species, making annotations even more difficult to obtain. Therefore, it is important to study which models and fine-tuning methods provide strong OOD robustness performance when trained with few ID images. We refer to this setting of fine-tuning a pre-trained model on low-shot ID images followed by evaluation on OOD images as the ``low-shot robustness'' setting (see Fig.~\ref{fig:teaser}).

From works that demonstrate robustness in the full-shot regime, we seem to arrive at the following conclusions for robustness to natural distribution shifts in the full-shot regime: (1) Amongst ImageNet pre-trained initializations, SSL ViTs are more robust than their supervised and CNN counterparts, with the more recent ones being better \cite{he2022masked, assran2022masked}. (2) Even without additional robustness interventions (i.e. methods to improve robustness), pre-trained models on large external datasets such as CLIP \cite{radford2021learning} provide superior robustness \cite{wortsman2022robust}. (3) Such models when combined with state-of-the-art robustness interventions lead to significant robustness improvements on several datasets \cite{kumar2022fine, wortsman2022robust, wortsman2022model}. In this paper, we question to what extent these conclusions hold true when the amount of training data is not as high. 

Overall, we perform the first in-depth study of robustness to various natural distribution shifts in different low-shot regimes: spanning datasets, architectures, pre-trained initializations, and state-of-the-art robustness interventions.
Through our experiments, we aim to answer the following key questions:

\begin{itemize}[leftmargin=*]

    \item[] \textbf{Q1.} \textit{For ImageNet pre-trained models, what kind of pre-training strategies and architectures are most effective for robustness in low-shot regimes?} \\
    \textbf{A}: Self-supervised ViTs generally perform better than CNNs and the supervised counterparts (where applicable) on both ID and OOD shifts, but no single initialization or model size works better across datasets.
    \begin{itemize}
        \item[•] For ImageNet and iWildCam \cite{beery2020iwildcam} datasets, MSN ViT \cite{assran2022masked} performs better than other models on OOD shifts, however a smaller model size (ViTS-16) works better for iWildCam but not for ImageNet.

        \item[•] For Camleyon \cite{bandi2018detection} dataset which is non object-centric, DINO ViTS-16 \cite{caron2021emerging} outperforms other models including DINO ViTB-16 and MSN ViTS-16 on both ID and OOD shifts.
    \end{itemize}

    \item[] \textbf{Q2.} \textit{Do models pre-trained on large external datasets, such as CLIP, provide superior robustness compared to ImageNet pre-trained ones on different datasets?} \\
    \textbf{A:} While we generally conform with the findings of recent works \cite{kumar2022fine, wortsman2022robust, wortsman2022model} and find that models such as CLIP \cite{radford2021learning} provide superior robustness on ImageNet and in full-shot regimes, we find that ImageNet pre-trained models can be better on other datasets such as iWildCam and Camelyon in the low-shot regimes.
    \begin{itemize}
        \item[•] Comparing ViTB-16 architecture on these datasets, DINO initialization outperforms CLIP (zero-shot or otherwise) and ImageNet-21k \cite{ridnik2021imagenet} supervised ViT on both ID and OOD shifts.
        \item[•] ImageNet supervised ViT \cite{touvron2021training} significantly outperforms ImageNet-21k supervised ViT on OOD shifts.
    \end{itemize}

    \item[] \textbf{Q3.} \textit{When using robustness interventions, does better robustness in the full-shot regime also imply better robustness in the low-shot regimes?} \\
    \textbf{A:} Not always. We find that depending on the initialization, existing interventions can fail to improve robustness in the full-shot regime or in some of the low-shot regimes for datasets other than ImageNet.
    \begin{itemize}
        \item[•] On iWildCam, interventions often fail to improve robustness with MSN ViTB-16 in the full-shot regime. On the other hand, only WiSE-FT \cite{wortsman2022robust} significantly improves robustness with CLIP ViTB-16 in both the full and low-shot regimes.
        \item[•] On Camelyon, while interventions often improve robustness in the full-shot regime for both MSN and CLIP ViTB-16,
        they fail to do so either in \textit{extreme} ($\sim 3000$ images) or in \textit{moderate} ($\sim 15000$ images) low-shot regimes, except WiSE-FT with CLIP.
    \end{itemize}
\end{itemize}

As highlighted by our findings, conventional wisdom for robustness to natural distribution shifts in the full-shot regime might not apply in the low-shot regimes, and should be seen as an important challenge for future work.

\section{Related Work}
\label{sec:related}
\par \noindent
\textbf{Robustness studies.} Real-world models may encounter and struggle to generalize on data distributions different than the ones used for training \cite{10.5555/1462129, torralba2011unbiased}. Previous works have studied such generalization capability of models under synthetic \cite{madry2017towards, tramer2017ensemble, geirhos2018generalisation, eykholt2018robust, alcorn2019strike, hendrycks2019benchmarking} and natural distribution shifts \cite{recht2019imagenet, barbu2019objectnet, wang2019learning, hendrycks2021natural, hendrycks2021many, koh2021wilds}. Researchers have also looked at the effect of architecture, i.e. CNNs and ViTs on robustness to different kinds of shifts \cite{naseer2021intriguing, bai2021transformers, zhou2022understanding, wang2022can} and distortion robustness of several models in comparison to humans \cite{geirhos2021partial}. In particular, \cite{taori2020measuring} performs a large-scale study of several supervised models and finds that interventions used for synthetic shifts offer little to no robustness gains for natural distribution shifts. On the other hand, accuracy under natural distribution shifts can often be reliably estimated by the in-distribution accuracy \cite{yadav2019cold, miller2020effect, taori2020measuring, miller2021accuracy} except for some shifts \cite{d2020underspecification, teney2022id}. Crucially, these works perform evaluations after training on fully labelled datasets with hundreds of thousands or millions of images which can be out of reach for most practitioners. 
While some recent works \cite{Azizi_2021_ICCV, wiles2022a, wenzel2022assaying} attempt to study the impact of training data amount on out-of-distribution robustness, they do not adopt the recent pre-training strategies \cite{radford2021learning, caron2021emerging, assran2022masked} and fine-tuning techniques \cite{kumar2022fine, wortsman2022robust, wortsman2022model} that have led to unprecedented robustness gains.
We therefore adopt such methods and perform experiments in the low-shot regime to observe its impact on robustness to various natural distribution shifts.

\par \noindent
\textbf{Self-supervised learning.} Researchers have shown self-supervised learning (SSL) to be better or on-par with supervised learning for pre-training deep networks for various downstream tasks \cite{he2020momentum, chen2020simple, caron2021emerging, ericsson2021well, he2022masked, assran2022masked} and we refer the reader to \cite{jing2020self, liu2021self} for thorough literature reviews. Recent methods that leverage ViTs \cite{he2022masked, assran2022masked} demonstrate superior robustness to some natural distribution shifts \cite{wang2019learning, hendrycks2021natural, hendrycks2021many} compared to previous state-of-the-art methods without additional interventions. However, such evaluations are performed only after fine-tuning on full ImageNet and whether the trend holds for other datasets and in different low-shot regimes remains an open question. We aim to address this gap in our work by evaluating some of the most recent SSL ViTs on a variety of datasets and distribution shifts, also comparing with CNNs and the supervised counterparts.

\par \noindent
\textbf{Few-shot learning.} Few-shot learning aims to generalize to novel classes from a few samples belonging to these classes \cite{lake2011one, lake2015human, vinyals2016matching}. While meta-learning based approaches used to be popular on standard benchmarks \cite{finn2017model, snell2017prototypical, sung2018learning, lee2019meta}, a growing wave of research showed that simpler transfer learning-based approaches can also achieve competitive performance \cite{chen2019closer, tian2020rethinking}. Recently, \cite{guo2020broader} conforms with this finding on the more challenging cross-domain few-shot learning (CD-FSL) scenario where the source and novel classes belong to different domains. Since then, works often perform SSL pre-training on the source data followed by low-shot fine-tuning on the few examples of novel classes \cite{ericsson2021well, zhang2022well, das2022confess}.
However, unlike the (cross-domain) few-shot scenario, we use the target or out-of-distribution (OOD) data only for evaluation purposes similar to most other robustness studies. Nonetheless, we use the classifiers adopted in \cite{guo2020broader} and present their detailed comparison on different datasets and associated design choices in section \ref{sec:appx_ls_ft} of appendix.
We discuss other related works that are either not applicable or already described in our experiments in appendix Sec.~\ref{sec:appx_rel_works}.

\section{Preliminaries: Robustness Metrics}
\label{sec:prelim}
Although using out-of-distribution (OOD) shifts to measure absolute performance can suggest robustness, it overlooks the in-domain (ID) performance of a model.
As pointed in~\cite{taori2020measuring}, two models with similar OOD performance can have vastly different ID performances. A better definition of robustness should consider the OOD performance beyond what is expected from achieving some level of ID performance. Therefore, to measure robustness, in addition to absolute performance comparison we also adopt the \textit{effective} and \textit{relative} robustness framework used in previous works \cite{recht2019imagenet, taori2020measuring, wortsman2022robust}. We now describe these metrics in detail.

Key to measuring effective robustness is establishing an expected baseline OOD accuracy given some ID accuracy $x$. This is established by computing a log-linear fit $\beta(x)$ over ID and OOD accuracies, i.e. $acc^s_{id}$ and $acc^s_{ood}$ respectively, for a set of standard models $f^{s}_1, f^{s}_2, \dots f^{s}_n$ as:
\setlength\abovedisplayskip{2pt}
\setlength\belowdisplayskip{2pt}
\begin{equation}
    \beta(x) = \sigma(w \text{ logit}(x) + b)
    \label{eq:logit}
\end{equation}

where $\text{logit}(x) = \ln{\frac{1}{1 - x}}$ and $\sigma$ is the inverse of the logit function. In practice, $\beta(x)$ is obtained by mapping each point $(x, y) \rightarrow (\text{logit}(x), \text{logit}(y))$ and solving linear regression. This can be visualized by plotting ($acc^{s}_{id}$, $acc^{s}_{ood}$) on a scatter plot with the $x$ and $y$ axes denoting ID and OOD accuracies respectively.

Once obtained, effective robustness of an ``intervention''\footnote{For models pre-trained on large external datasets such as CLIP~\cite{radford2021learning}, it's unclear what datasets are considered in or out-of-distribution, so we exclude it from the standard set of models and treat it as an intervention.} $r$ applied on the model $f^{s}$, i.e. $f^r = (acc^{r}_{id}, acc^{r}_{ood})$ can be expressed as:
\setlength\abovedisplayskip{2pt}
\setlength\belowdisplayskip{2pt}
\begin{equation}
    \rho(f^{r}) = acc^{r}_{ood} - \beta(acc^{r}_{id})
    \label{eq:eff_rob}
\end{equation}
which outlines if the intervention leads to OOD accuracy beyond what is expected from having a higher ID accuracy.

While effective robustness is important, it is not enough to provide a comprehensive evaluation of models, especially in the low-shot regimes. An ``intervention'' on a model may result in high positive $\rho(f^{r})$, indicating effective robustness, but it could also decrease both ID and OOD accuracies which is not desirable. Thus, in addition to effective robustness, we measure relative robustness by assessing the impact of an intervention on OOD accuracy as:
\setlength\abovedisplayskip{2pt}
\setlength\belowdisplayskip{2pt}
\begin{equation}
    \tau(f^{r}) = acc^{r}_{ood} - acc^{s}_{ood}
    \label{eq:rel_rob}
\end{equation}

Following~\cite{taori2020measuring}, an intervention $r$ is said to improve the robustness of a model $f^{s}$ only when it exhibits both effective and relative robustness, that is, $\rho(f^{r}) > 0$ and $\tau(f^{r}) > 0$. However, our experiments indicate that interventions frequently lack simultaneous effective and relative robustness across various low-shot regimes. For simplicity, we refer to $\rho(f^{r})$ as $\rho$ and $\tau(f^{r})$ as $\tau$.

\section{Experimental Setting}
\label{sec:setting}
Following prior work, we assume full label-space overlap and study image classification under natural distribution shifts~\cite{radford2021learning, wortsman2022robust}. 
Additionally, we refer to low-shot as $10^{3} - 10^{4}$ images, as also shown in Fig.~\ref{fig:teaser} and table \ref{tab:ls_regimes}.
We describe our experimental setting with the associated design choices and justifications in this section.

\subsection{Datasets and Low-Shot Regimes}
\label{sec:dsets_and_regimes}
\begin{table}[t]
\centering
\setlength{\tabcolsep}{2.5pt}
\resizebox{0.9\columnwidth}{!}{
\begin{tabular}{l*{2}ccc}
\toprule
\multirow{2}{*}{\textbf{Dataset}} & \multicolumn{3}{c}{\textbf{Low-Shot Regimes (Imgs / Class)}}  \\
& Extreme & Moderate & High  \\
\hline
\texttt{1} ImageNet~\cite{deng2009imagenet} & $1$ & $5$  & $\sim13$  \\
\texttt{2} iWildCam~\cite{beery2020iwildcam} & $1$-$480$ & $1$-$4802$ & $1$-$9604$  \\
\texttt{3} Camelyon~\cite{bandi2018detection} & $1500$  & $7500$  & $15000$   \\
\bottomrule
\end{tabular}}
\caption{\small\textbf{Different Low-Shot Regimes.} We consider low-shot regimes with similar number of images for different datasets and describe them in more detail in section \ref{sec:dsets_and_regimes}.}
\label{tab:ls_regimes}
\vspace{-15pt}
\end{table}

Prior studies~\cite{taori2020measuring,miller2021accuracy}
have observed a linear trend for certain supervised models on ImageNet~\cite{deng2009imagenet} and iWildCam \cite{beery2020iwildcam} datasets after applying the logit function (see Eq.~\ref{eq:logit}), while contrasting evidence has been reported for other datasets, such as Camelyon~\cite{bandi2018detection}, in~\cite{teney2022id}. To obtain a comprehensive view of robustness in low-shot regimes, where a strong correlation between in-domain (ID) and out-of-distribution (OOD) performances may or may not exist, we conduct experiments on all three datasets.

\par\noindent
\textbf{ImageNet \& Distribution Shifts.}
ImageNet (IN1k)~\cite{deng2009imagenet} is
an extensive dataset for image recognition that consists of objects and scenes belonging to one of the 1000 classes. For training, we use the subsets with 1, 5, and $\sim$13 images per class (see table~\ref{tab:ls_regimes} row 1) used by~\cite{assran2022masked} for comparison of
self-supervised methods based on in-domain (ID) accuracy.
We use the IN1k validation split for model validation based on top-1 accuracy.\footnote{While the optimal model checkpoint for ID performance might not be so for OOD performance, it is a widely adopted practice~\cite{taori2020measuring, kumar2022fine} that allows for a fair comparison across different methods.} For testing, we report the average top-1 accuracy on the following $5$ natural distribution shifts:

\par\noindent
\underline{ImageNet-R (IN-R)}~\cite{hendrycks2021many} has 200 classes in common with IN1k and rendition images such as sculptures and paintings.
\par\noindent
\underline{ImageNet-S (IN-S)}~\cite{wang2019learning} consists of around 50,000 images of sketches, similar to the size of IN1k's validation set.
\par\noindent
\underline{ImageNet-A (IN-A)}~\cite{hendrycks2021natural} has 200 classes in common with IN1k and images that are classified  incorrectly by a supervised ResNet-50 (RN50) \cite{he2016deep} trained on IN1k.
\par\noindent
\underline{ImageNet-v2 (IN-v2)}~\cite{recht2019imagenet} consists of similar images as in IN1k's test set but from a different distribution.
\par\noindent
\underline{ObjectNet (ON)}~\cite{barbu2019objectnet} has 113 common classes
with IN1k and images that vary in rotation, background, and viewpoint.

\begin{figure}[t]
\centering
\includegraphics[width=\columnwidth]{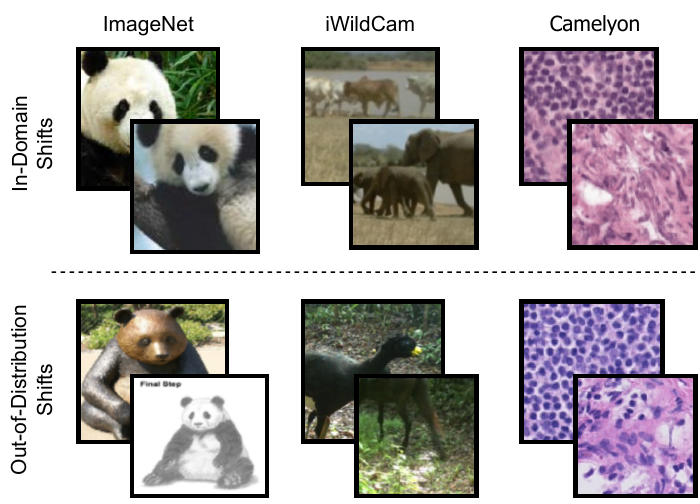}
\caption{\small\textbf{Datasets \& Distribution Shifts.} We show some sample images from ImageNet \cite{deng2009imagenet}, iWildCam \cite{beery2020iwildcam}, and Camelyon \cite{bandi2018detection} datasets and associated distribution shifts \cite{wang2019learning, hendrycks2021many}.}
\label{fig:datasets}
\end{figure}

\par\noindent
\textbf{iWildCam.} The iWildCam~\cite{beery2020iwildcam} dataset 
comprises images of $182$ animal species captured by various cameras traps, which are treated as different distributions.
We use the WILDS benchmark \cite{koh2021wilds} and manually curate low-shot subsets from \texttt{train} shift for training which has $129809$ images, \texttt{val-id} shift for validation which has $7314$ images, and \texttt{val-ood} shift for testing which has $14961$ images. 
Since these sets have an imbalanced class distribution, We sample images from the different classes in 1\%, 10\%, and 20\% ratios from the \texttt{train} shift, ensuring that each class has at least one image. This results in low-shot subsets with $1370$, $12973$, and $25931$ images, respectively (see table~\ref{tab:ls_regimes} row 2).
Additionally, we report average per-class accuracy for both validation and testing.

\par\noindent
\textbf{Camelyon.} The Camelyon~\cite{beery2020iwildcam} dataset consists of $96\times96$ histopathological images that may or may not contain tumor tissue,
resulting in $2$ classes. These scans are sourced from different hospitals that are considered different distributions. We again use the WILDS benchmark~\cite{koh2021wilds} and manually curate low-shot subsets from \texttt{train} shift for training which has $302436$ images, \texttt{val-id} shift for validation which has $33560$ images, and \texttt{val-ood} shift for evaluation which has $34904$ images.\footnote{We emphasize that the \texttt{val-ood} shifts are used only for evaluation. While \texttt{test-ood} shifts are also available in WILDS benchmark~\cite{koh2021wilds}, they have similar creation processes but larger number of images than the \texttt{val-ood} shifts, so we opt for the latter due to limited compute.} The shifts are well-balanced, so we create subsets containing $1500$, $7500$, and $15000$ images per class (see table~\ref{tab:ls_regimes} row 3). We report the average per-class accuracy for validation and testing and find that it is within 1 percentage point of top-1 accuracy.

\subsection{Standard Models}
\label{sec:std_models}

Recall that to establish a baseline out-of-distribution accuracy $\beta(x)$ for a given in-domain accuracy $x$, fitting Eq.~\ref{eq:logit} for a set of ``standard'' models is required. We consider a standard
set of ImageNet (IN1k) pre-trained models for this purpose, which are not subjected to additional robustness interventions or pre-training data. The selection of these models is based on their low-shot ID performance comparison on IN1k~\cite{assran2022hidden} or average performance on various downstream tasks~\cite{ericsson2021well}. We show the architectures used in our experiments ordered by the number of parameters (i.e.~size) below. More detailed comparison is also shown in table \ref{tab:appx_clip_params} in appendix.
$$\text{ViTS-16} \approx \text{RN50} < \text{ViTB-16} \approx \text{RN50w2} < \text{ViTL-16}$$

\par\noindent
\textbf{Self-supervised models.} We include the following self-supervised (SSL) models for our experiments.
\par\noindent
\underline{SwAV}~\cite{caron2020unsupervised}: SwAV is a SSL method for pre-training CNNs by predicting cluster assignments for different augmented views of an image and enforcing consistency between them. We use the RN50 and RN50w2 checkpoints for all datasets.
\par\noindent
\underline{DINO}~\cite{caron2021emerging}: DINO
self-trains a student network to match the feature embeddings of augmented local and global views of an image to that of a teacher network which sees only the global view. We use the RN50, ViTS-16, and ViTB-16 checkpoints for all datasets.
\par\noindent
\underline{MSN}~\cite{assran2022masked}: MSN
matches the predicted cluster assignments for masked and unmasked augmented views of an image and performs well on low-shot ID evaluation on ImageNet. We use the ViTS-16 and ViTB-16 checkpoints for all datasets and ViTL-16 checkpoint for ImageNet.

\par\noindent
\textbf{Supervised models.} For datasets other than ImageNet, we additionally include \underline{DEIT}~\cite{touvron2021training} (ViTS-16, ViTB-16) and \underline{supervised ResNet-50} from PyTorch~\cite{paszke2019pytorch}. Note that for ImageNet, these models violate the ``low-shot'' condition as they have already been trained with the full labelled dataset.

\par\noindent
\textbf{Fitting Standard Models.} To obtain the parameters ($w$ and $b$) of the log-linear curve for $\beta(x)$ in Eq.~\ref{eq:logit} for a dataset, we first train individual models from the standard set on different low-shot subsets and full-shot subset (details in Sec.~\ref{sec:ls_ft}). Fine-tuning and subset details are provided in appendix, Sec.~\ref{sec:appx_std_sub}. We then evaluate the trained models on both ID validation and OOD test shifts, out of which \textit{only} the former is used for hyperparameter tuning.
In case of multiple OOD test shifts, we calculate the average OOD performance following previous work~\cite{wortsman2022robust}. We assess the quality of the curve fit via mean absolute error ($\text{MAE}$) and coefficient of determination ($R^2$) of the curve on these data points, as shown in table~\ref{tab:quality}.
Finally, the curve $\beta(x)$ is used to calculate effective robustness of an intervention using Eq.~\ref{eq:eff_rob}.

\par\noindent
\textbf{Low-Shot Training.}
\label{sec:ls_ft}
For low-shot training with the standard models,~we freeze the pre-trained models and train a classifier on top with the available training data.
We compare the following classifiers based on prior work in cross-domain few-shot learning \cite{guo2020broader} -- Logistic Regression~\cite{tian2020rethinking}, Mean-Centroid Classifier~\cite{snell2017prototypical}, and Baseline++~\cite{chen2019closer} -- and select the best-performing one for each dataset.
While Logistic Regression performs better or on-par on both ID and OOD shifts for ImageNet and iWildCam, Baseline++ performs better on Camelyon. We provide this comparison and more details in section \ref{sec:appx_ls_ft} of appendix.

\subsection{Robustness Interventions}

We consider some of the most recent methods for improving robustness to natural distribution shifts and models pre-trained on large external datasets as robustness interventions (see Sec.~\ref{sec:related}). We briefly summarize them below:

\begin{itemize}[leftmargin=0pt,label={}]
\setlength\itemsep{0em}
\item \underline{LP-FT}~\cite{kumar2022fine}: LP-FT follows a two-stage strategy of first fine-tuning only the randomly initialized linear head followed by fine-tuning the entire model end-to-end on fully labelled datasets.
\item \underline{CLIP}~\cite{radford2021learning}: CLIP is a vision-language model that is pre-trained on a large number of ($\sim$ 400M) image-text pairs. It shows strong zero-shot performance on several datasets and is often used as the de-facto initialization by several works \cite{kumar2022fine, wortsman2022model, wortsman2022robust}.
\item \underline{WiSE-FT}~\cite{wortsman2022robust}: WiSE-FT applies a weight-space ensemble between a zero-shot model such as CLIP and this model fine-tuned on fully labelled datasets. For IN1k pre-trained models, we ensemble between the weights of linear-probed (LP) and LP-FT checkpoints due to the absence of a zero-shot head. We use $\alpha = 0.5$ unless mentioned otherwise.
\item \underline{Model Soups}~\cite{wortsman2022model}: Model Soups uses a weight-space ensemble of several models that are trained with a different epochs, learning rates, weight decay, label smoothing \cite{szegedy2016rethinking}, mixup \cite{zhang2017mixup}, and RandAugment \cite{cubuk2020randaugment}. Due to limited compute and the scale of experiments, we use a greedy soup with 9 models and again use linear-probing for the head initialization. We follow the paper for hyperparameter values.
\item \underline{RobustViT}~\cite{chefer2022optimizing}: RobustViT first uses an unsupervised object localization method such as TokenCut \cite{wang2022self} to dump offline segmentation maps. It then optimizes a supervised ViT's saliency maps \cite{chefer2021generic} to resemble these offline segmentation maps while maintaining its classification accuracy.
\end{itemize}

For a uniform comparison across datasets, we apply the relevant interventions on MSN ViTB-16 and use it as the reference model for computing effective and relative robustness (see Sec.~\ref{sec:prelim}).
Additionally, we include CLIP with LP-FT, WiSE-FT, and Model Soups as interventions, based on their reported performances \cite{kumar2022fine, wortsman2022robust, wortsman2022model} and strong performance on ID and OOD shifts in our experiments.
Despite being amenable to low-shot training, it remains challenging to implement RobustViT on non-object centric datasets such as Camelyon due to its requirement of offline segmentation maps.
We provide details on the hyperparameter choices for every intervention in section \ref{sec:appx_interventions} of appendix.

\section{Results}
\label{sec:results}
We now present findings for $3$ key questions 
from Sec.~\ref{sec:intro} -- (1) among ImageNet pre-trained models, which ones are more robust in low-shot regimes (see table~\ref{tab:ls_regimes})
(2) how do they compare with models pre-trained on larger datasets and (3) do robustness interventions help in the low-shot regimes.

\subsection{Comparing ImageNet Pre-trained Models}
\begin{figure}
    \centering
    {{\includegraphics[width=8cm]{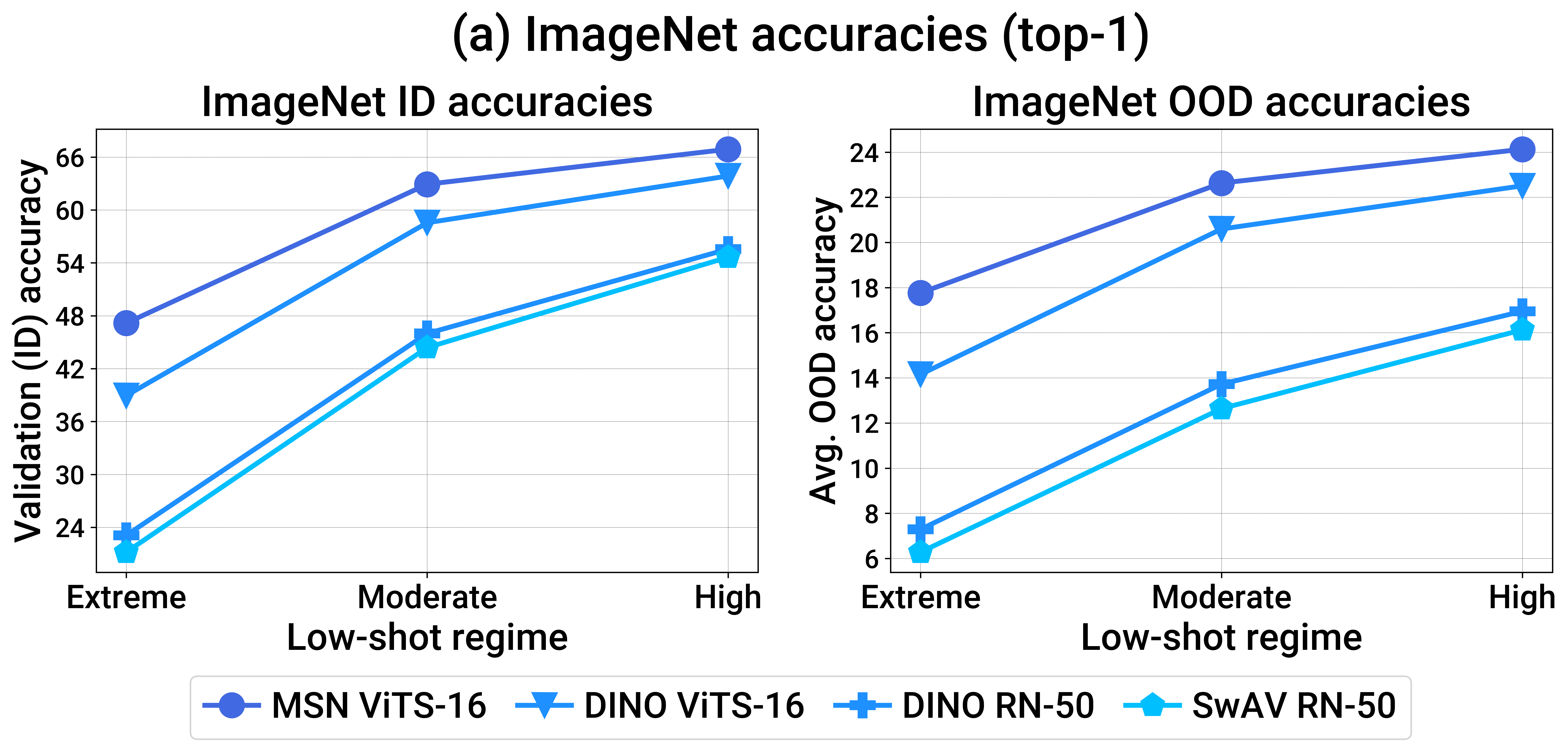}}}\\\vspace{1mm}
    {{\includegraphics[width=8cm]{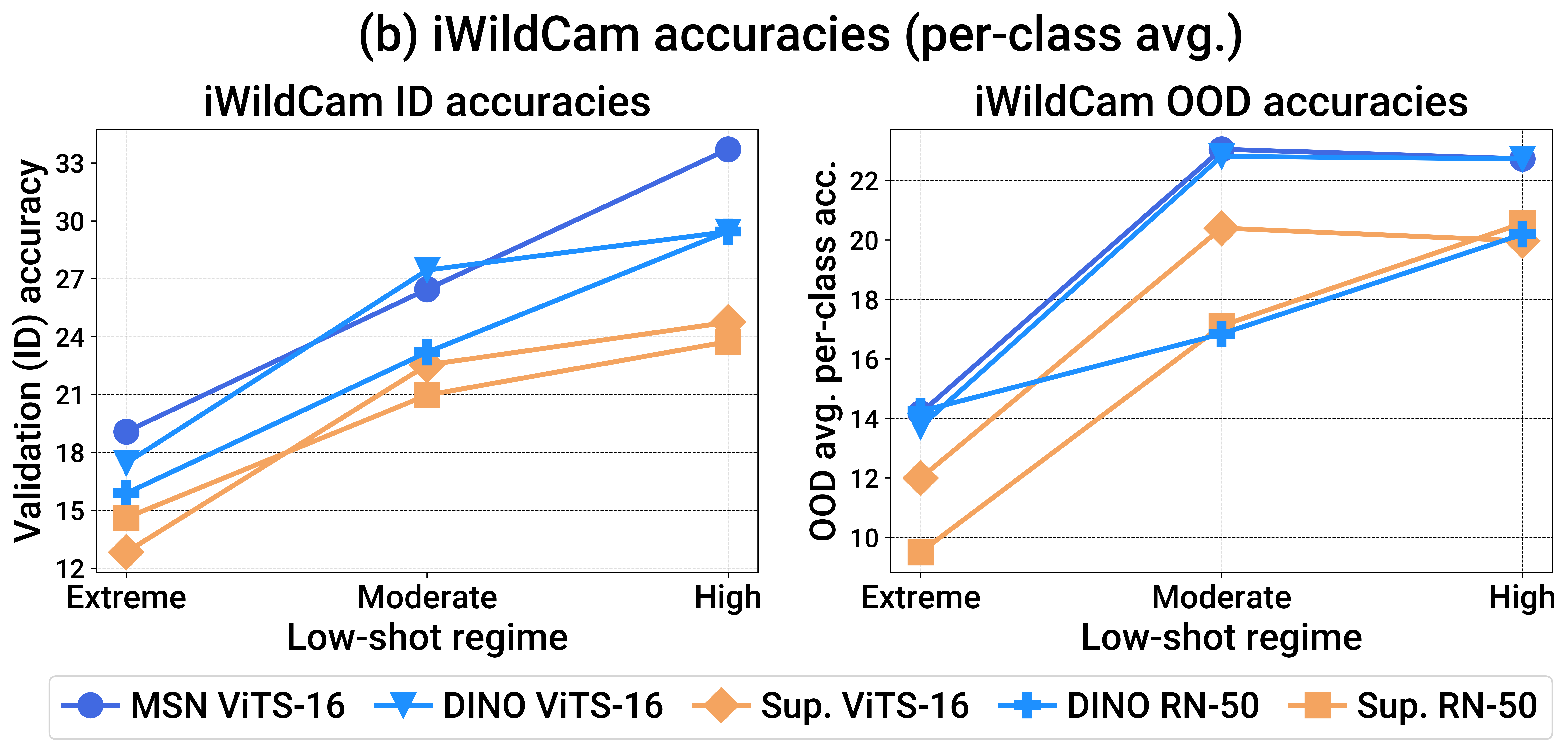}}}\\\vspace{1mm}
    {{\includegraphics[width=8cm]{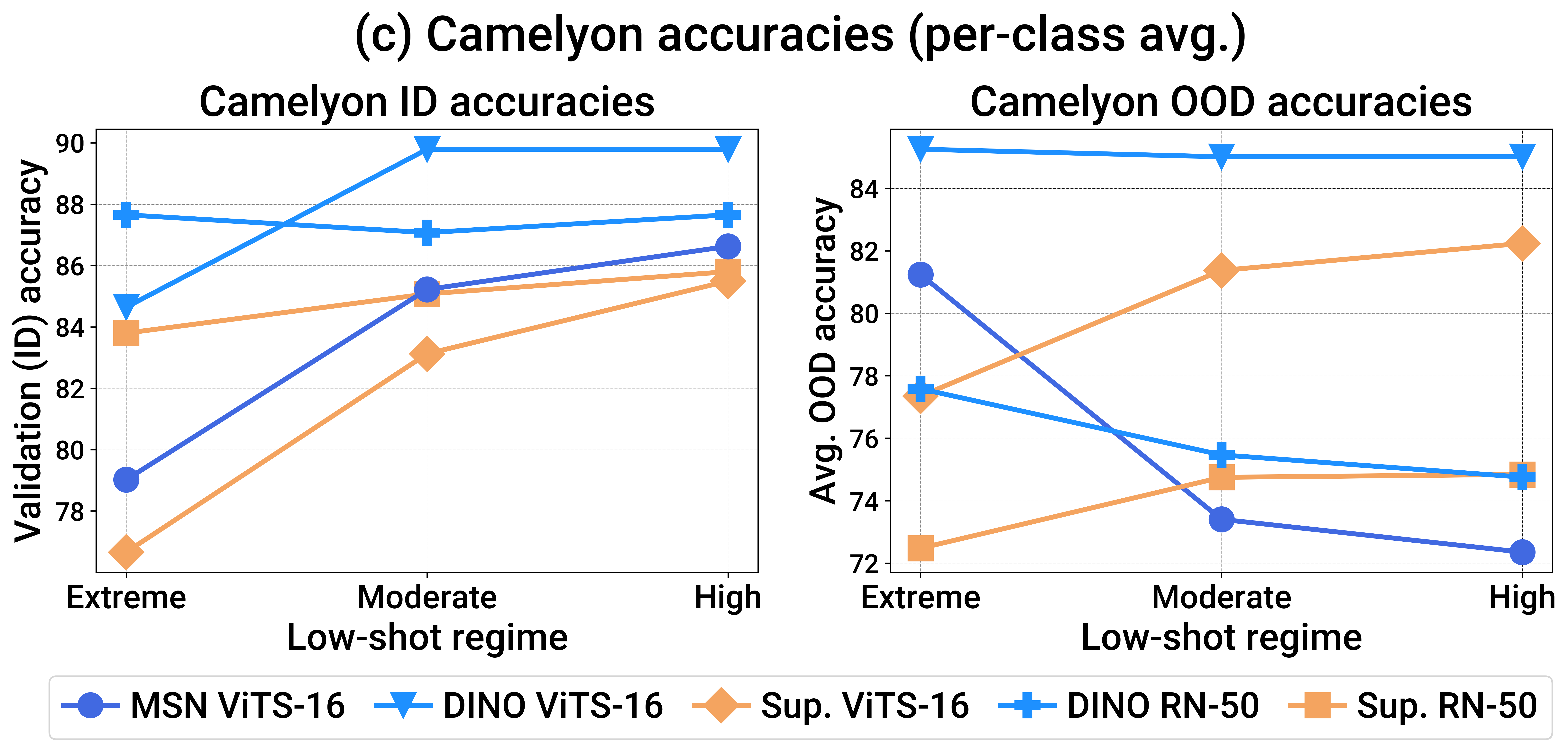}}}
    \caption{\small\textbf{Comparison of ImageNet pre-trained architectures and initializations.} 
    With similar number of parameters, self-supervised (SSL) ViTs generally perform better on both ID and OOD shifts compared to SSL CNNs and the supervised counterparts where applicable.
    }
    \label{fig:arch}
\end{figure}

We compare ImageNet pre-trained models with similar number of parameters (ViTS-16 and RN50) on the basis of absolute ID and OOD performances in Fig.~\ref{fig:arch}. For a uniform comparison, we randomly initialize the classifier head (see Sec.~\ref{sec:std_models}) and use the same hyperparameters for all models. It can be seen that self-supervised (SSL) ViTs often perform better than SSL CNNs on ImageNet and supervised ViTs and CNNs on iWildCam and Camelyon datasets.

However, the best initialization and model size varies for each dataset as shown in table \ref{tab:in_ssl_vits}. For a concise comparison, we show the average ID and OOD performances across different low-shot regimes. MSN ViTB-16 outperforms DINO ViTB-16 and MSN ViTS-16 on ImageNet, but not on iWildCam where MSN ViTS-16 performs better on OOD shift. Similarly, DINO ViTS-16 performs better than other models on both ID and OOD shifts on Camelyon.

Thus, while SSL ViTs perform better than SSL CNNs and the supervised counterparts (where applicable) on both ID and OOD shifts in the low-shot regimes, no single initialization or model size performs the best across datasets.

\begin{table}[t]
\centering
\setlength{\tabcolsep}{2.5pt}
\resizebox{\columnwidth}{!}{
\begin{tabular}{@{} l rr rr rr @{} }
\toprule
 & \multicolumn{2}{c}{ImageNet} & \multicolumn{2}{c}{iWildCam} & \multicolumn{2}{c}{Camelyon} \\
 \cmidrule(lr){2-3} \cmidrule(lr){4-5} \cmidrule(lr){6-7}
 &  ID & OOD & ID & OOD & ID & OOD \\
\midrule
  \texttt{1} MSN ViTS-16~\cite{assran2022masked} & 58.99 & 21.51 & 26.41 & \textbf{19.99} & 83.62 & 75.67 \\
  \texttt{2} DINO ViTS-16~\cite{caron2021emerging} & 53.78 & 19.09 & 24.78 & 19.75 & \textbf{88.08} & \textbf{85.09} \\
  \texttt{3} MSN ViTB-16 ~\cite{assran2022masked} & \textbf{61.40} & \textbf{22.81} & 24.78 & 19.65 & 86.40 & 78.84 \\
  \texttt{4} DINO ViTB-16 \cite{caron2021emerging} & 56.72 & 21.98 & \textbf{27.40} & 19.82 & 86.93 & 84.33 \\
\bottomrule
\end{tabular}
}
\caption{\small\textbf{Comparison of ImageNet pre-trained self-supervised ViTs.} On average across low-shot regimes, no single self-supervised initialization or model size outperforms others on ID and OOD shifts across datasets.}
\label{tab:in_ssl_vits}
\end{table}

\subsection{Pre-training Data Scale and Strategy}

\begin{table}[t]
\centering
\setlength{\tabcolsep}{2.5pt}
\resizebox{\columnwidth}{!}{
\begin{tabular}{@{} l rr rr rr @{} }
\toprule
 & \multicolumn{2}{c}{ImageNet} & \multicolumn{2}{c}{iWildCam} & \multicolumn{2}{c}{Camelyon} \\
 \cmidrule(lr){2-3} \cmidrule(lr){4-5} \cmidrule(lr){6-7}
 & ID & OOD & ID & OOD & ID & OOD \\
\midrule
  \texttt{1} CLIP zero shot \cite{radford2021learning, wortsman2022robust} & \textbf{67.93} & \textbf{57.37} & 9.67 & 16.82 & 50.48 & 51.55 \\
  \texttt{2} CLIP \cite{radford2021learning} & 50.8 & 27.50 & 23.75 & 19.10 & 84.9 & 77.3 \\
  \texttt{3} Supervised (IN21k) \cite{dosovitskiy2020image} & N/A & N/A & 16.84 & 16.90 & 85.18 & 81.07 \\
  \texttt{4} Supervised (IN1k) \cite{touvron2021training} & N/A & N/A & 22.27 & 18.57 & 83.35 & 83.24 \\
  \texttt{5} MSN (IN1k) \cite{assran2022masked} & 61.40 & 22.81 & 24.78 & 19.65 & 86.40 & 78.84 \\
  \texttt{6} DINO (IN1k) \cite{caron2021emerging} & 56.72 & 21.98 & \textbf{27.40} & \textbf{19.82} & \textbf{86.93} & \textbf{84.33} \\
\bottomrule
\end{tabular}
}
\caption{\small\textbf{Comparison between ViTs pre-trained on different datasets.} On average across low-shot regimes, ImageNet (IN) pre-trained SSL ViT's such as DINO are worse than CLIP on ImageNet. However, it performs much better than CLIP and IN-21k supervised ViT on iWildCam and Camelyon datasets.
}
\label{tab:data_vs_strat}
\end{table}

We question whether models pre-trained on large external datasets provide superior robustness over ImageNet (IN1k) pre-trained ones in the low-shot regimes without additional interventions. We compare CLIP ViT and a supervised ViT pre-trained on ImageNet-21k (IN21k) \cite{dosovitskiy2020image, ridnik2021imagenet} with IN1k pre-trained ViT's. We use the ViTB-16 architecture with the same classifiers described in Sec.~\ref{sec:ls_ft}.

We again compare the absolute performance on ID and OOD shifts on average across low-shot regimes. As with the IN1k supervised models, IN21k supervised ViT violates the ``low-shot'' premise so we don't use it on ImageNet. For CLIP zero-shot results, we match the implementation of \cite{wortsman2022robust} and provide additional details in appendix, Sec.~\ref{sec:appx_ls_ft}.

As shown in table~\ref{tab:data_vs_strat}, CLIP's zero-shot performance on ID and OOD shifts on ImageNet is significantly better than both CLIP and IN1k pre-trained models.
However, CLIP (zero-shot or otherwise) performs worse than IN1k pre-trained models on iWildCam and Camelyon, on which DINO performs better than other models. IN21k supervised ViT often performs significantly worse than IN1k supervised ViT on these datasets, especially on OOD shifts.

Thus, IN1k pre-trained models can perform better on both ID and OOD shifts than the models pre-trained on large external datasets in low-shot regimes, on datasets such as iWildCam and Camelyon.

\subsection{Effect of Robustness Interventions}

We question the extent to which existing robustness interventions improve robustness in the low-shot regimes, and we examine how the trend compares with the full-shot regime. We present the dataset-wise observations below.

\begin{figure*}
    \centering
    \includegraphics[width=\linewidth]{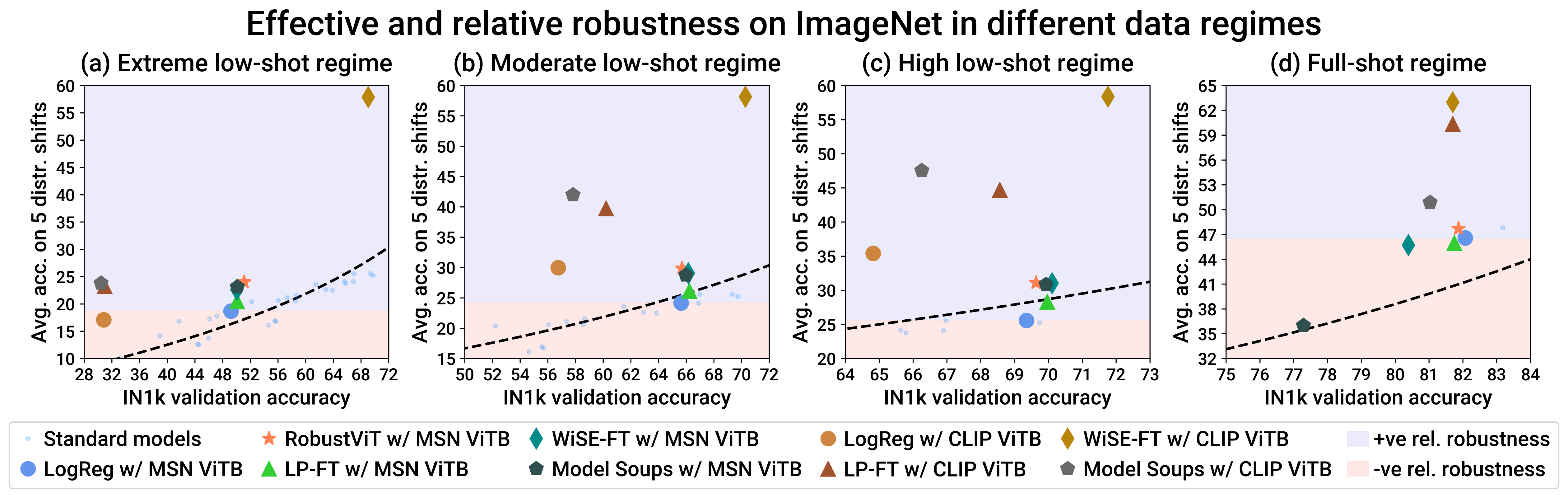}
    \caption{\small\textbf{Effect of robustness interventions on ImageNet.} Plots (a), (b), and (c) show performance of interventions in low-shot regimes (see table \ref{tab:ls_regimes}).
    Plot (d) shows performance of interventions in the full-shot regime.
    Interventions located above the line ($\rho > 0$) and in the blue region ($\tau > 0$) are said to improve robustness (see Sec. \ref{sec:prelim}). 
    Interventions largely improve robustness in the low-shot regimes with MSN ViTB-16, and
    in all data regimes when coupled with CLIP ViTB-16.
    }
    \label{fig:ir_inet}
\end{figure*}

\begin{figure*}
    \centering
    \includegraphics[width=\linewidth]{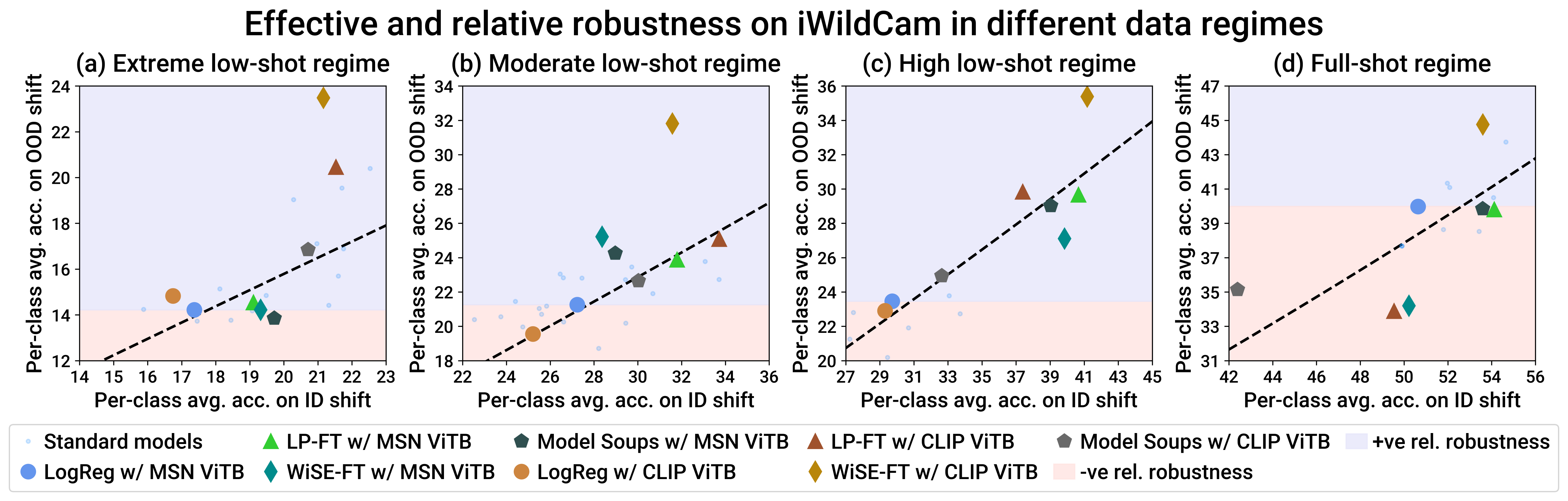}
    \caption{\small\textbf{Effect of robustness interventions on iWildCam.} Interventions often fail to improve robustness in both the full and low-shot regimes with MSN ViTB-16. Only WiSE-FT with CLIP ViTB-16 improves robustness in all data regimes.
    }
    \label{fig:ir_iwc}
\end{figure*}

\begin{figure*}
    \centering
    \includegraphics[width=\linewidth]{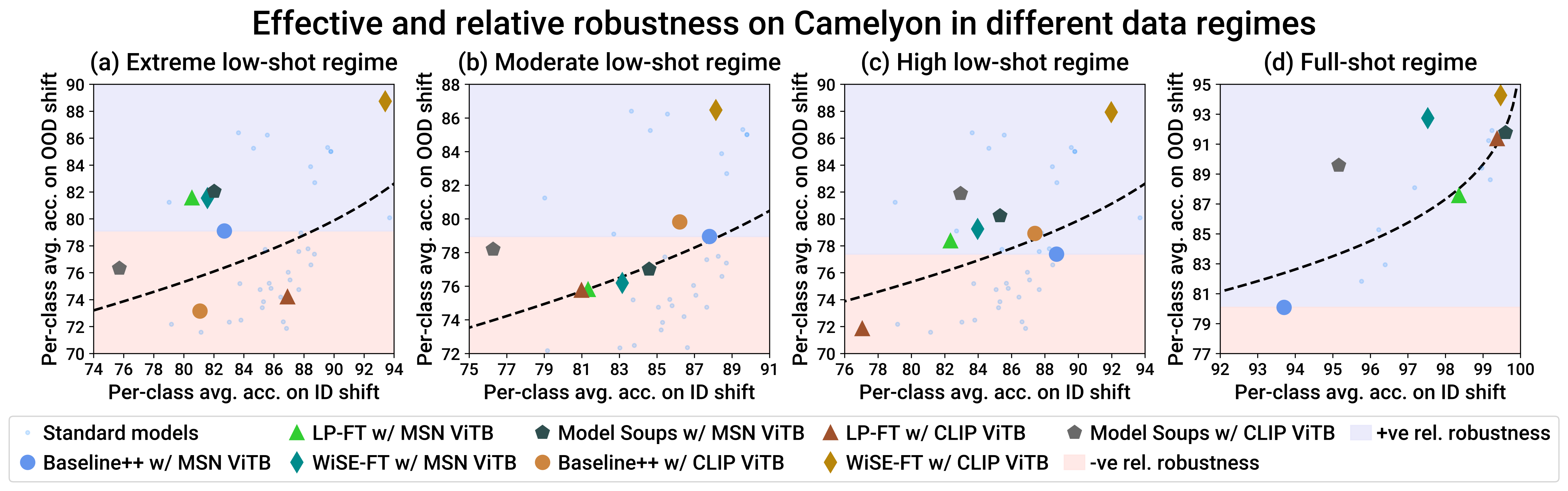}
    \caption{\small\textbf{Effect of robustness interventions on Camelyon.} Interventions often improve robustness in the full-shot regime with both MSN and CLIP ViTB-16 but fail to do so in \textit{extreme} or \textit{moderate} low-shot regimes, except WiSE-FT with CLIP ViTB-16.
    }
    \label{fig:ir_cmlyn}
\end{figure*}

\textbf{ImageNet.} We show the results of this experiment in Fig.~\ref{fig:ir_inet}. With MSN, interventions are largely effectively and relatively robust in the different low-shot regimes, except LP-FT in the \textit{high} low-shot regime. While the interventions are also effectively robust in the full-shot regime, they are often not relatively robust,
except RobustViT which improves robustness in all regimes.

When coupled with CLIP, Model Soups and WiSE-FT also become effectively and relatively robust in all data regimes with the latter providing largest robustness improvements. Zero-shot CLIP also improves robustness significantly in low-shot regimes (see table \ref{tab:ir_full_low}), suggesting that not using the limited training data is a better approach. However, we find that it is not the case on other datasets.

\textbf{iWildCam.} We show the results of this experiment in Fig.~\ref{fig:ir_iwc}. With MSN, interventions are often relatively but not effectively robust in the low-shot regimes and neither effectively nor relatively robust in the full-shot regime. Unlike ImageNet, CLIP's zero-shot performance is quite poor (see table \ref{tab:ir_full_low}) and WiSE-FT with CLIP is the only intervention which improves robustness in all data regimes.

\begin{table}[t]
\centering
\begin{tabular}{l*{2}c}
\toprule
Dataset & $\text{MAE} \downarrow$ & $R^2 \uparrow$ \\
\hline
\texttt{1} ImageNet \cite{deng2009imagenet} & 1.75 & 0.94 \\
\texttt{2} iWildCam \cite{beery2020iwildcam} & 1.50 & 0.96 \\
\texttt{3} Camelyon \cite{bandi2018detection} & 3.44 & 0.50 \\
\bottomrule
\end{tabular}
\caption{\small\textbf{Quality of curve fit.} 
Curve $\beta(x)$ fit on the accuracies of standard models (see Sec.~\ref{sec:std_models}) 
leads to a relatively higher $\text{MAE}$ and lower $R^2$ on Camelyon, 
indicating the poor quality of fit.}
\label{tab:quality}
\vspace{-15pt}
\end{table}

\textbf{Camelyon.} We show the results of this experiment in Fig.~\ref{fig:ir_cmlyn}. Note that for Camelyon, the quality of curve $\beta(x)$ fit with the ID and OOD accuracies of \textit{standard} models is relatively low compared to other datasets as shown in table \ref{tab:quality}, in which case relative robustness should be prioritized since it doesn't rely on the quality of the fit.
While interventions improve relative robustness in the full-shot regime with MSN, they fail to do so in the \textit{moderate} low-shot regime. Similarly, interventions improve robustness in the full-shot regime with CLIP, but LP-FT fails to be relatively robust in the \textit{moderate} low-shot regime whereas Model Soups fails to be relatively robust in both the \textit{extreme} and \textit{moderate} low-shot regimes. 
Only WiSE-FT ($\alpha = 1$) with CLIP improves robustness in all data regimes.
CLIP's zero-shot performance is near random as shown in table \ref{tab:data_vs_strat}.

We show the effective and relative robustness of the interventions in the full-shot regimes in table \ref{tab:ir_full_low}. To complement our findings, we also highlight the interventions which significantly\footnote{We use the standard deviation of residuals obtained after fitting $\beta(x)$ to determine significance, and provide more details in appendix, Sec. \ref{sec:appx_significance}.} 
improve robustness across both the full-shot regime and majority of low-shot regimes for each dataset.
We see that (1) most interventions significantly improve robustness on ImageNet but not on other datasets and (2) no intervention significantly improves robustness across datasets and data regimes, except WiSE-FT with CLIP.

\begin{table}[t]
\centering
\setlength{\tabcolsep}{1.5pt}
\resizebox{\columnwidth}{!}{
\begin{tabular}{@{} l rr rr rr @{} }
\toprule
 & \multicolumn{2}{c}{\textbf{ImageNet}} & \multicolumn{2}{c}{\textbf{iWildCam}} & \multicolumn{2}{c}{\textbf{Camelyon}} \\ 
 \cmidrule(lr){2-3} \cmidrule(lr){4-5} \cmidrule(lr){6-7} 
 & \multicolumn{1}{c}{$\rho \uparrow$} & \multicolumn{1}{c}{$\tau \uparrow$} & \multicolumn{1}{c}{$\rho \uparrow$} & \multicolumn{1}{c}{$\tau \uparrow$} & \multicolumn{1}{c}{$\rho \uparrow$} & \multicolumn{1}{c}{$\tau \uparrow$} \\
\midrule
\textbf{Full-Shot Regime} \\
\midrule
  \texttt{1} LP-FT \cite{kumar2022fine} & \textcolor{gray}{5.16} & \textcolor{gray}{-0.61} & \textcolor{gray}{-1.41} & \textcolor{gray}{-0.17} & \textcolor{gray}{-0.45} & \textcolor{gray}{7.48} \\
  \texttt{2} \quad + CLIP & \sigrob{19.60$^*$} & \sigrob{$\text{13.77}^*$} & \textcolor{gray}{-3.60} & \textcolor{gray}{-6.09} & 0.37 & 11.28 \\
  \texttt{3} WiSE-FT \cite{wortsman2022robust} & \textcolor{gray}{6.66} & \textcolor{gray}{-0.86} & \textcolor{gray}{-3.84} & \textcolor{gray}{-5.87} & 6.22 & 12.66 \\
  \texttt{4} \quad + CLIP & \sigrob{$\text{22.24}^*$} & \sigrob{$\text{16.41}^*$} & \sigrob{3.98} & \sigrob{4.78} & \sigrob{2.85} & \sigrob{14.18} \\
  \texttt{5} Model Soups \cite{wortsman2022robust} & \textcolor{gray}{0.53} & \textcolor{gray}{-10.58} & \textcolor{gray}{-0.93} & \textcolor{gray}{-0.14} & \textcolor{gray}{-0.35} & \textcolor{gray}{11.68} \\
  \texttt{6} \quad + CLIP & \sigrob{$\text{11.00}^{\dagger}$} & \sigrob{$\text{4.29}^{\dagger}$} & \textcolor{gray}{3.20} & \textcolor{gray}{-4.84} & 5.93 & 9.50 \\
  \texttt{7} RobustViT \cite{chefer2022optimizing} & \sigrob{6.73} & \sigrob{1.13} & \textcolor{gray}{N/A} & \textcolor{gray}{N/A} & \textcolor{gray}{N/A} & \textcolor{gray}{N/A} \\
  \texttt{8} CLIP zero-shot \cite{radford2021learning, wortsman2022robust} & \sigrob{30.28} & \sigrob{10.79} & \textcolor{gray}{8.46} & \textcolor{gray}{-23.167} & \textcolor{gray}{-14.63} & \textcolor{gray}{-28.54} \\
\bottomrule
\end{tabular}
}
\caption{\small\textbf{Robustness intervention comparison.} The table shows effective ($\rho$) and relative ($\tau$) robustness of different interventions in the full-shot regime.
$*$ and $\dagger$ denote numbers obtained from papers for ViTB-16 and ViTB-32 architecture respectively. 
Interventions that do not improve robustness in the full-shot regime are shown in \textcolor{gray}{gray}, while interventions that do so are shown in black. Interventions that significantly improve robustness in \textit{both} the full-shot regime and majority of low-shot regimes are highlighted in \sigrob{blue} for each dataset.
Robustness results for the low-shot regimes (as shown in Fig.~\ref{fig:ir_inet},~\ref{fig:ir_iwc}, and~\ref{fig:ir_cmlyn}) are also provided in the appendix.
Most interventions significantly improve robustness on ImageNet but not on other datasets, except WiSE-FT with CLIP.
}
\label{tab:ir_full_low}
\vspace{-10pt}
\end{table}

We also measure the statistical significance of our results by obtaining the mean and standard deviation across $2$ different runs and show them in table \ref{tab:appx_perf_var} of appendix. We observe that the best performing interventions such as WiSE-FT with CLIP also exhibit small (within $2$ pp) variance.

\par \noindent
\textbf{Limitations.} We note that there are limitations to our study.
First, we were unable to theoretically analyze our results due to the vast and empirical nature of our study. Recent works \cite{park2022vision, park2023selfsupervised} demonstrate the data specificity of ViTs and the global semantic invariance of SSL approaches such as DINO, which can be helpful for this purpose.
Second, we were unable to observe the effects of in-domain SSL pre-training on datasets other than ImageNet.
Recent work \cite{assran2022hidden} has also shown that the current objectives of self-supervised methods such as MSN and DINO might not be suitable for class-imbalanced datasets (e.g.~iWildCam).
Third, while we incorporate different kinds of augmentations and loss functions as a part of interventions such as Model Soups,
singly analyzing their effect on robustness in low-shot regimes remains an avenue for future work.

\section{Conclusion}
We conclude our study of low-shot robustness to several natural distribution shifts, which addresses the gap in the literature and marks the first in-depth study of its kind.
Taken together, our results demonstrate that: (1) Modern architectures (i.e.~ViT) and pre-training strategies (i.e.~self-supervised learning) lead to better robustness in low-shot regimes, but the best initialization and model size is dataset dependent. (2) Without additional interventions, large-scale vision-language pre-training can be underwhelming compared to ImageNet pre-trained models on datasets other than ImageNet.
(3) Robustness in the full-shot regime may not imply robustness in low-shot regimes on datasets other than ImageNet. While the performance of interventions is largely dependent on datasets and initializations, ensembling in weight-space seems promising to bridge this gap.
We hope that our study will motivate researchers to also focus on this problem of practical importance.

\section*{Acknowledgements}
\noindent
This work was supported in part by NASA ULI, NSF \#2144194, and ARL. We thank the anonymous reviewers for their thoughtful comments and suggestions. We also thank our internal reviewers Anisha Pal, Bharat Goyal, and Pratik Ramesh for their constructive suggestions on an early draft of this paper.

{\small
\bibliographystyle{ieeetr}
\bibliography{egpaper}
}

\newpage
\section{Training details}
\label{sec:appx_ft_setting}

\subsection{Full-shot fine-tuning} 
We follow MSN \cite{assran2022masked} for linear-probing and MAE \cite{he2022masked} for full fine-tuning of standard models (see Sec.~\ref{sec:appx_std_sub}) on ImageNet \cite{deng2009imagenet}. For iWildCam \cite{beery2020iwildcam} and Camelyon \cite{bandi2018detection} datasets, we follow the WILDS benchmark \cite{koh2021wilds} for fine-tuning design choices. We summarize some of these in table \ref{tab:appx_ft_choices}.

\begin{table}[t]
\centering
\begin{tabular}{@{} l r r r @{}}
\toprule
 & \multicolumn{1}{c}{ImageNet} & \multicolumn{1}{c}{iWildCam} & \multicolumn{1}{c}{Camelyon} \\
\midrule
Type & Linear & Full & Full \\
L2-normalization & True & False & False \\
Optimizer & SGD \cite{ruder2016overview} & Adam \cite{kingma2014adam} & SGD \cite{ruder2016overview} \\
Scheduler & Cosine & None & None \\
Epochs & 100 & 12 & 10 \\
Batch size & 128 & 16 & 32 \\
Learning rate & 6.4 & 0.00001 & 0.001 \\
Momentum & 0.9 & (0.9, 0.999) & 0.9 \\
Weight decay & 0 & 0 & 0.01 \\
\bottomrule
\end{tabular}
\caption{\textbf{Fine-tuning design choices.} We summarize some of the design choices for linear probing on ImageNet and full fine-tuning on other datasets, following \cite{assran2022masked} and \cite{koh2021wilds}.}
\label{tab:appx_ft_choices}
\end{table}

\subsection{Low-shot training} 
\label{sec:appx_ls_ft}

For low-shot training, we freeze the pre-trained models and train a classifier on top with the available training data. Based on the BS-CDFSL study \cite{guo2020broader}, we compare the following classifiers and use the best performing one in terms of in-domain (ID) performance for each dataset:
\begin{itemize}
    \setlength\itemsep{0em}
    \item Logistic Regression \cite{tian2020rethinking}: Linear head is applied on feature embeddings (optionally L2-normalized) and trained with a cross-entropy loss. We follow the implementation of MSN \cite{assran2022masked} which uses (\texttt{Resize, CenterCrop, Normalize}) augmentations and Cyanure \cite{mairal2019cyanure} package for training and evaluation.
    \item Mean-Centroid Classifier \cite{snell2017prototypical}: 
    Per-class cluster embeddings are obtained by averaging the feature embeddings of every image in the training data for that class. Then, predicted label for a test image is the corresponding label of the nearest (in terms of L2 distance) cluster center.
    \item Baseline++ \cite{chen2019closer}: 
    Also uses a linear head but the logits are obtained via cosine similarity between head weights and L2-normalized feature embeddings.
    We match their implementation and use (\texttt{RandomResizedCrop, ImageJitter, RandomHorizontalFlip, Normalize}) augmentations, and compare
    design choices in table~\ref{tab:appx_cf_choices}.
\end{itemize}

\begin{table}[t]
\centering
\begin{tabular}{@{} l r r @{}}
\toprule
 & \multicolumn{1}{c}{LogReg \cite{assran2022masked}} & \multicolumn{1}{c}{Baseline++ \cite{chen2019closer}} \\
\midrule
Normalization & Layer norm \cite{ba2016layer} & Weight norm \cite{salimans2016weight} \\
Optimizer & \texttt{auto} \cite{mairal2019cyanure} & SGD \cite{ruder2016overview} \\
Epochs & 300 & 100 \\
Learning rate & N/A & 0.01 \\
Batch size & 16 & 16 \\
Weight decay & 0.0025 & 0.001 \\
\bottomrule
\end{tabular}
\caption{\textbf{Classifier design choices.} We summarize some of the design choices for the different classifiers used for low-shot training. LogReg stands for Logistic Regression.}
\label{tab:appx_cf_choices}
\end{table}

We show their comparison with MSN ViTS-16 on different datasets in table~\ref{tab:appx_cf_results}. On average across low-shot regimes, Logistic Regression performs better on ID and OOD shifts on ImageNet, better on ID shift and on-par (within 1 \% point) on OOD shift on iWildCam. However, Baseline++ performs better on ID and OOD shifts on Camelyon.

\begin{table*}[t]
\centering
\label{results table}
\begin{tabular}{@{} l rr rr rr @{} }
\toprule
 & \multicolumn{2}{c}{ImageNet accs. (Top-1)} & \multicolumn{2}{c}{iWildCam accs. (Avg.)} & \multicolumn{2}{c}{Camelyon accs. (Avg.)} \\
 \cmidrule(lr){2-3} \cmidrule(lr){4-5} \cmidrule(lr){6-7} 
 & ID & OOD & ID & OOD & ID & OOD \\
\midrule
  Logistic Regression & \textbf{58.99} & \textbf{21.51} & \textbf{26.41} & 19.99 & 73.85 & 69.73 \\
  Mean Centroid Classifier & 57.46 & 20.5 & 24.33 & \textbf{20.72} & 81.12 & 70.26 \\
  Baseline++ & 48.6 & 21.10 & 17.74 & 14.62 & \textbf{83.62} & \textbf{75.66} \\
\bottomrule
\end{tabular}
\caption{\textbf{Classifier comparison across datasets.} We compare the $3$ classifiers -- Logistic Regression \cite{tian2020rethinking, assran2022masked}, Mean Centroid Classifier \cite{snell2017prototypical}, and Baseline++ \cite{chen2019closer} -- on average across low-shot regimes on different datasets with the MSN ViTS-16 model. Logistic Regression performs better on both ID and OOD shifts on ImageNet, better on ID shift and on-par on OOD shift on iWildCam. However, Baseline++ performs better on both ID and OOD shifts on Camelyon.
}

\label{tab:appx_cf_results}
\end{table*}

\par\noindent \\
\textbf{Additional details for CLIP \cite{radford2021learning}.} We use the ViTB-16 and RN50 models as they have the closest number of parameters to the different models under consideration as shown in table~\ref{tab:appx_clip_params}. As with the standard models, we freeze the pre-trained models and train the classifiers (Baseline++ for Camelyon, Logisitic Regression for others) with the available training data. We compare the average performance on the low-shot regimes (see table~\ref{tab:appx_std_model_regimes}) for these models in table~\ref{tab:appx_arch_clip}, and observe that ViTB-16 significantly outperforms RN50 on all datasets. Hence we use it for additional experiments with the robustness interventions.

For zero-shot results, we match the implementation of \cite{wortsman2022robust} who use a set of 80 and 2 prompts for ImageNet and iWildCam respectively. We use the prompt \texttt{"a photo of a <class> patch"} for Camelyon where \texttt{class $\in$ \{normal, tumor\}} following \cite{koh2021wilds, wortsman2022robust}. More specifically, we initialize the final classification layer of CLIP ViTB-16 with the zero-shot head constructed via these set of prompts. Following \cite{wortsman2022robust}, we also scale the head weights with CLIP's temperature parameter and L2-normalize its outputs before feeding them into the zero-shot head.
\begin{table*}[t]
\centering
\begin{tabular}{@{} l rr rr rr @{} }
\toprule
 & \multicolumn{2}{c}{ImageNet accs. (Top-1)} & \multicolumn{2}{c}{iWildCam accs. (Avg.)} & \multicolumn{2}{c}{Camelyon accs. (Avg.)} \\
 \cmidrule(lr){2-3} \cmidrule(lr){4-5} \cmidrule(lr){6-7} 
 & ID & OOD & ID & OOD & ID & OOD \\
\midrule
  CLIP ViTB-16 & \textbf{50.80} & \textbf{27.50} & \textbf{23.75} & \textbf{19.1} & \textbf{84.9} & \textbf{77.3} \\
  CLIP RN50 & 35.93 & 11.24 & 18.04 & 14.17 & 70.24 & 64.42 \\
\bottomrule
\end{tabular}
\caption{\textbf{Architecture comparison with CLIP \cite{radford2021learning}.} We compare the CLIP ViTB-16 architecture with the RN50 variant on average across low-shot regimes. ViTB-16 significantly outperforms RN50 on both ID and OOD shifts.
}
\label{tab:appx_arch_clip}
\end{table*}

\section{Standard models and subsets}
\label{sec:appx_std_sub}
For obtaining the log-linear curve $\beta(x)$, we use the following subsets and standard models, i.e. trained on ImageNet without additional robustness interventions:

\begin{table}[t]
\centering
\setlength{\tabcolsep}{2.5pt}
\resizebox{0.9\columnwidth}{!}{
\begin{tabular}{l*{2}cccc}
\toprule
\multirow{2}{*}{\textbf{Dataset}} & \multicolumn{4}{c}{\textbf{Low-Shot Regimes (Imgs / Class)}}  \\
& Extreme & Low & Moderate & High  \\
\hline
ImageNet~\cite{deng2009imagenet} & $1$ & $2$ & $5$ & $\sim13$  \\
iWildCam~\cite{beery2020iwildcam} & $1$-$480$ & $1$-$2401$ & $1$-$4802$ & $1$-$9604$  \\
Camelyon~\cite{bandi2018detection} & $1500$ & $3000$ & $7500$  & $15000$   \\
\bottomrule
\end{tabular}}
\caption{\small\textbf{Different Low-Shot Regimes.} We use the subsets described in this table for fitting the curve $\beta(x)$ (see Eq.~\ref{eq:appx_logit}). Note that only the \textit{extreme}, \textit{moderate}, and \textit{high} low-shot regimes are used in the rest of our experiments for simplicity.}
\label{tab:appx_std_model_regimes}
\end{table}

\textbf{ImageNet.} We use the $1$, $2$, $5$, and $\sim${$13$} images per class subsets provided by \cite{assran2022masked} for low-shot training. The initializations and model sizes used are:
\begin{itemize}
    \setlength\itemsep{0em}
    \item \underline{MSN} \cite{assran2022masked}: ViTS-16, ViTB-16, and ViTL-16
    \item \underline{DINO} \cite{caron2021emerging}: RN50, ViTS-16, and ViTB-16
    \item \underline{SwAV} \cite{caron2020unsupervised}: RN50 and RN50w2
\end{itemize}

Here, we only use the MSN ViTB-16 and DINO ViTB-16 models for the full-shot regime due to limited compute.

\textbf{iWildCam.} We create subsets with images in $1\%$, $5\%$, $10\%$, and $20\%$ ratio of the original \texttt{train} shift in WILDS \cite{koh2021wilds} benchmark while ensuring that each of the $182$ classes have at least one image. These subsets have $1,370$, $6,510$, $12,973$, and $25,931$ images respectively. The standard models used for this dataset in all data regimes are:
\begin{itemize}
    \setlength\itemsep{0em}
    \item \underline{MSN} \cite{assran2022masked}: ViTS-16 and ViTB-16
    \item \underline{DINO} \cite{caron2021emerging}: RN50, ViTS-16, and ViTB-16
    \item \underline{SwAV} \cite{caron2020unsupervised}: RN50 and RN50w2
    \item \underline{DEIT} \cite{touvron2021training}: ViTS-16 and ViTB-16
    \item \underline{Supervised RN50} \cite{paszke2019pytorch}
\end{itemize}

\textbf{Camelyon.} We create subsets with $1,500$, $3,000$, $7,500$, and $15,000$ images per class from \texttt{train} shift in WILDS \cite{koh2021wilds} benchmark for each of the $2$ classes. We use the same set of models as iWildCam for this dataset.
\begin{table}[t]
\centering
\begin{tabular}{l*{2}c}
\toprule
Model & Parameters  \\
\hline
RN50 \cite{paszke2019pytorch} & 23,508,032 \\
CLIP RN50 \cite{radford2021learning} & 38,316,896 \\
RN50w2 \cite{touvron2021training} & 93,907,072 \\
\midrule
ViTS-16 \cite{touvron2021training} & 21,664,896 \\
ViTB-16 \cite{touvron2021training} & 85,797,120 \\
ViTB-16 (IN21k) \cite{dosovitskiy2020image} & 86,389,248 \\
CLIP ViTB-16 \cite{radford2021learning} & 57,844,224 \\
ViTL-16 \cite{touvron2021training} & 303,299,584 \\
\bottomrule
\end{tabular}
\caption{\textbf{Parameter comparison.} Comparison of number of trainable parameters (without classifier) between different models in the same architecture family.}
\label{tab:appx_clip_params}
\vspace{-10pt}
\end{table}

We summarize these subsets for all datasets in table~\ref{tab:appx_std_model_regimes}. For simplicity, we only use the \textit{extreme}, \textit{moderate}, and \textit{high} low-shot regimes for the rest of our experiments. Our code and low-shot subsets are publicly available at 
\href{https://github.com/Aaditya-Singh/Low-Shot-Robustness/}{this url}.

\section{Robustness interventions}
\label{sec:appx_interventions}

We now describe the design choices and hyperparameters used for all interventions. Our general strategy is to use the model checkpoint which (a) trains to near completion, i.e a training accuracy of $98\%-100\%$ and (b) leads to the highest in-distribution (ID) validation accuracy. Following \cite{wortsman2022robust} who observe that models with similar ID performance can have vastly different OOD performance, we generally use the smallest learning rate that meets these criteria.

\subsection{LP-FT \cite{kumar2022fine}}
\label{sec:appx_lp-ft}
LP-FT adopts a two-stage strategy of freezing the pre-trained model and training a randomly initialized head, followed by full fine-tuning the entire model. We mostly follow table \ref{tab:appx_ft_choices} for the values of different hyperparameters except for the ones described below.

\textbf{ImageNet.} We use the linear probing (LP) hyperparameters provided by MSN \cite{assran2022masked} as also shown in table~\ref{tab:appx_ft_choices}. For full fine-tuning in the full-shot regime, we use the MAE codebase \cite{he2022masked} and fine-tune for $20$ epochs.
In the low-shot regimes, we use the hyperparameters shown in table~\ref{tab:appx_ft_choices} except a learning rate of $0.0001$ for LP-FT following \cite{kumar2022fine}.

\textbf{iWildCam.} We do a grid search over the number of epochs ($ep$), learning rate ($lr$), and weight decay ($wd$) for linear probing and find a combination of $(120, 0.001, 0.001)$ to work well across models and data regimes.
For ImageNet pre-trained models, we linear probe for $240$ epochs in low-shot regimes and use a combination of $(ep = 12, lr = 0.00001, wd = 0)$ for full fine-tuning. As the intervention is primarily meant for CLIP, we do a grid search over ($ep \in \{12, 24\}, lr \in \{0.00001, 0.000001\}, wd \in \{0.001, 0.01, 0.0\}$) and select the checkpoint with the best ID validation performance.

\textbf{Camelyon.} We do a grid search over the number of epochs, learning rate, and weight decay for linear probing and find a combination of $(ep = 20, lr = 0.001, wd = 0.001)$ to work well across models and data regimes.
For ImageNet pre-trained models, we find a combination of $(ep = 12, lr = 0.0001, wd = 0.01)$ to work well. As for CLIP, we do a grid search over ($ep \in \{10, 20\}, lr \in \{0.00001, 0.000001\}, wd \in \{0.001, 0.01, 0.0\}$) and select the checkpoint with the best ID validation performance.

\subsection{WiSE-FT \cite{wortsman2022robust}}
WiSE-FT ensembles between the weights of a zero-shot model such as CLIP and this model fine-tuned in the full-shot regime. The method has a mixing coefficient $\alpha$ which determines the relative weight assigned to the fine-tuned model with respect to the zero-shot model, i.e.~$\theta = (1 - \alpha) \cdot \theta_0 + \alpha \cdot \theta_1$ where $\theta, \theta_0, \theta_1$ refer to the weights of the model after ensembling, the zero-shot model, and the fine-tuned model respectively.

 Since ImageNet pre-trained models such as MSN don't have a zero-shot head, we use LP and LP-FT models (see Sec.~\ref{sec:appx_lp-ft}) for the weight space ensemble.
 For CLIP, we ensemble between the weights of the pre-trained model with a zero-shot head (see Sec.~\ref{sec:appx_ls_ft}) and this model fine-tuned fully.
 For ImageNet, we use the same hyperparameters described in section~\ref{sec:appx_ft_setting} except a learning rate of $0.00001$ in the low-shot regimes for better ID performance. Otherwise, we perform a grid search over hyperparameters as for LP-FT (see Sec.~\ref{sec:appx_lp-ft}) and select the best ID validation checkpoint. 

Following \cite{wortsman2022robust}, we use an $\alpha = 0.5$ unless mentioned otherwise. With CLIP on Camelyon, we search over $\alpha \in \{0, 0.5, 1\}$ and report the $\alpha$ which achieves the highest ID validation performance, i.e.~$\alpha = 1$. We show this comparison with along the OOD performances in table \ref{tab:appx_alpha_wiseft}.

\begin{table}[t]
\centering
\begin{tabular}{@{}lrc@{}}
\toprule
    & \multicolumn{2}{c}{Camelyon accs. (Avg.)} \\ \cmidrule(l){2-3} 
    & \multicolumn{1}{c}{ID} & \multicolumn{1}{c}{OOD} \\
\midrule
\textbf{Full-Shot} \\ 
\midrule
$\alpha=0$ & 50.48 & 51.55 \\
$\alpha=0.5$ & 75.68 & 70.60 \\
$\alpha=1$ & \textbf{99.47} & \textbf{94.27} \\
\midrule
\textbf{(Average) Low-Shot} \\ 
\midrule
$\alpha=0$ & 50.48 & 51.55 \\
$\alpha=0.5$ & 61.33 & 59.98 \\
$\alpha=1$ & \textbf{91.18} & \textbf{87.71} \\
\bottomrule
\end{tabular}
\caption{\textbf{WiSE-FT \cite{wortsman2022robust} $\alpha$ comparison.} We compare the ID and OOD performances of WiSE-FT with CLIP for different $\alpha$ values on Camelyon dataset. $\alpha = 1$ results in significantly better performance across data regimes.
}
\label{tab:appx_alpha_wiseft}
\end{table}

\subsection{Model Soups \cite{wortsman2022model}}
Model Soups performs a weight space ensemble with several models that are fine-tuned with different set of augmentations and optimizer configurations. The associated hyperparameters for each model in the soup are chosen randomly, and the value ranges following \cite{wortsman2022model} are shown in table~\ref{tab:appx_modelsoups_hparams}. Due to limited compute, we use a greedy soup 
\footnote{We find that this version of the soup performed substantially better than the uniform soup on iWildCam \cite{beery2020iwildcam} dataset in all data regimes.}
of $9$ models for our experiments in which a fine-tuned model is greedily added to the soup only if its ID performance is enhanced after adding the current model to the soup.

\begin{table}[t]
\centering
\begin{tabular}{@{}lcl@{}}
\toprule
                & Value Range &  \\ 
\midrule
Epochs & [$4,16$] &  \\
Learning Rate & [$10^{-4}$,$10^{-6}$] &  \\
Weight Decay & [$10^{-0.2}$,$10^{-4}$] &  \\
Label Smoothing \cite{szegedy2016rethinking} & [0,0.25] &  \\
Mixup \cite{zhang2017mixup} & [0,0.9] & \\
RandAug \cite{cubuk2020randaugment} $M$ & [0,20] & \\
RandAug \cite{cubuk2020randaugment} $N$ & [0,2] &  \\
\bottomrule
\end{tabular}
\caption{\textbf{Model Soups \cite{wortsman2022model} hyperparameters.} Value ranges for each hyperparameter used in the random search.
}
\label{tab:appx_modelsoups_hparams}
\end{table}

\subsection{RobustViT \cite{chefer2022optimizing}}

RobustViT uses an unsupervised localization method such as TokenCut \cite{wang2022self} to dump offline segmentation maps and then optimizes a supervised ViT's saliency maps \cite{chefer2021generic} to resemble the offline ones while maintaining its classification accuracy to its improve robustness on the OOD shifts for ImageNet \cite{recht2019imagenet, barbu2019objectnet, wang2019learning, hendrycks2021natural, hendrycks2021many}.

First, we use TokenCut to dump the segmentation maps for each of the images in the $1$, $5$ and $\sim${$13$} images per class subsets. Then, we follow the original authors' implementation for fine-tuning with the proposed augmentations, losses, and hyperparameters. However, we find that these lead to poor performance for self-supervised (SSL) ViTs such as MSN ViTB-16, likely due to the absence of a classification head for such models.

Thus, we first perform a linear probing step with the hyperparameters used for LP-FT and described in section \ref{sec:appx_lp-ft} for $50$ epochs, and then perform the proposed fine-tuning with the default hyperparameters. For the full-shot regime, we use our fine-tuned model checkpoint (see Sec.~\ref{sec:appx_ft_setting}) and directly perform the proposed fine-tuning step with the $2$ images per class subset, as it's close to the number of images used in \cite{chefer2022optimizing}. We find this strategy to work well which significantly improves robustness of MSN ViTB-16 across data regimes, as shown in table \ref{tab:ir_full_low} in the main paper.

For datasets other than ImageNet and especially Camelyon which is non object-centric, we note that the method remains challenging to implement primarily due to the need of offline segmentation maps.

\section{Measuring significance for robustness.}
\label{sec:appx_significance}

The effective ($\rho$) and relative ($\tau$) robustness metrics \cite{recht2019imagenet, taori2020measuring, wortsman2022robust} can be used to determine whether a robustness intervention $r$ applied on a standard model $f^s$, i.e.~$f^r$ improves robustness or not (see Sec.~\ref{sec:prelim} in main paper). However, these metrics don't inform whether an intervention which improves robustness does so \textit{significantly} or not. An intervention $r$ can technically improve robustness but barely so, \textcolor{black}{i.e. $\rho, \tau \rightarrow 0^{+}$}. Also, the quality of curve fit $\beta(x)$ could be poor (table \ref{tab:quality} in main paper) due to which a simple strategy such as $\rho > \rho_{0} \text{ and } \tau > \tau_{0}$ for some $\rho_{0} \text{ and } \tau_{0}$ might not be suitable. Therefore, we use the standard deviation of the points used to fit the curve $\beta(x)$ for measuring significance.

Specifically, given a set $S$ of in-domain (ID) and out-of-distribution (OOD) accuracies of $n$ standard models, i.e.
\setlength\abovedisplayskip{3pt}
\setlength\belowdisplayskip{3pt}
\begin{equation}
    S = \{(acc^{k}_{id}, acc^{k}_{ood}) \; \forall \, k \, \in [n]\}
    \label{eq:appx_std_accs}
\end{equation}

Recall that log-linear curve $\beta(x)$ is defined as:
\setlength\abovedisplayskip{3pt}
\setlength\belowdisplayskip{3pt}
\begin{equation}
    \beta(x) = \sigma(w \text{ logit}(x) + b)
    \label{eq:appx_logit}
\end{equation}

where $\text{logit}(x) = \ln{\frac{1}{1 - x}}$ and $\sigma$ is the inverse of the logit function. Each point in set $S$ is mapped by $\text{logit}(x)$ and $\beta(x)$ is obtained by using the mapped points to solve linear regression. Next, we obtain the set of residuals $R$ as:
\setlength\abovedisplayskip{3pt}
\setlength\belowdisplayskip{3pt}
\begin{equation}
    R = \{\text{logit}(acc^{k}_{ood}) - (w \text{ logit}(acc^{k}_{id}) + b) \; \forall \, k \in [n]\}
    \label{eq:appx_residuals}
\end{equation}

We then compute the standard deviation $d$ of the set of residuals $R$ as:
\setlength\abovedisplayskip{3pt}
\setlength\belowdisplayskip{3pt}
\begin{equation}
    d = \sqrt{\frac{\sum_{k=1}^{n}R_k^2}{n - 2}}
    \label{eq:appx_std_dev}
\end{equation}

Next, we define $\beta_{\lambda}(x)$ which can be thought of as a \textcolor{black}{shifted version} of $\beta(x)$, as:
\begin{equation}
    \beta_{\lambda}(x) = \sigma(w\,x + b + \lambda \,d)
    \label{eq:appx_beta_lambda}
\end{equation}

Finally, we say that an intervention $r$ applied on a standard model $f^s$, i.e. $f^r = (acc^{r}_{id}, acc^{r}_{ood})$ significantly improves robustness if both the following conditions hold:
\begin{align}
    acc^{r}_{ood} > \beta_{\lambda}(acc^{r}_{id})
    \label{eq:appx_sig_rob1} \\
    acc^{r}_{ood} > acc^{s}_{ood} + \gamma
    \label{eq:appx_sig_rob2}
\end{align}

\textcolor{black}{where $\lambda \text { and } \gamma$ can be arbitrary, but we opt for $\lambda = 1 \text{ and } \gamma = 0$ for a milder definition of significance.}
We provide the values for $w$, $b$, and $d$ to define $\beta(x)$ and $\beta_{\lambda}(x)$ for each dataset in table~\ref{tab:appx_beta_lambda_params}.

Intuitively, we ask whether the intervention provides an OOD accuracy that is (1) one standard deviation beyond the OOD accuracy that can be expected from its ID accuracy after logit transform and (2) better than the OOD accuracy of the standard model without the intervention (or $\tau > 0$).
Across multiple data regimes, an intervention is said to significantly improve robustness if it does so (Eq.~\ref{eq:appx_sig_rob1} and \ref{eq:appx_sig_rob2}) in the full-shot regime and on majority of low-shot regimes.

\begin{table*}[t]
\centering
\begin{tabular}{@{} l rr rr rr @{} }
\toprule
 & \multicolumn{2}{c}{\textbf{ImageNet}} & \multicolumn{2}{c}{\textbf{iWildCam}} & \multicolumn{2}{c}{\textbf{Camelyon}} \\ 
 \cmidrule(lr){2-3} \cmidrule(lr){4-5} \cmidrule(lr){6-7} 
 & \multicolumn{1}{c}{$\rho \uparrow$} & \multicolumn{1}{c}{$\tau \uparrow$} & \multicolumn{1}{c}{$\rho \uparrow$} & \multicolumn{1}{c}{$\tau \uparrow$} & \multicolumn{1}{c}{$\rho \uparrow$} & \multicolumn{1}{c}{$\tau \uparrow$} \\
\midrule
\textbf{Full-Shot Regime} \\
\midrule
  \texttt{1} LP-FT \cite{kumar2022fine} & \textcolor{gray}{5.16} & \textcolor{gray}{-0.61} & \textcolor{gray}{-1.41} & \textcolor{gray}{-0.17} & \textcolor{gray}{-0.45} & \textcolor{gray}{7.48} \\
  \texttt{2} \quad + CLIP & \sigrob{19.60$^*$} & \sigrob{$\text{13.77}^*$} & \textcolor{gray}{-3.60} & \textcolor{gray}{-6.09} & 0.37 & 11.28 \\
  \texttt{3} WiSE-FT \cite{wortsman2022robust} & \textcolor{gray}{6.66} & \textcolor{gray}{-0.86} & \textcolor{gray}{-3.84} & \textcolor{gray}{-5.87} & 6.22 & 12.66 \\
  \texttt{4} \quad + CLIP & \sigrob{$\text{22.24}^*$} & \sigrob{$\text{16.41}^*$} & \sigrob{3.98} & \sigrob{4.78} & \sigrob{2.85} & \sigrob{14.18} \\
  \texttt{5} Model Soups \cite{wortsman2022robust} & \textcolor{gray}{0.53} & \textcolor{gray}{-10.58} & \textcolor{gray}{-0.93} & \textcolor{gray}{-0.14} & \textcolor{gray}{-0.35} & \textcolor{gray}{11.68} \\
  \texttt{6} \quad + CLIP & \sigrob{$\text{11.00}^{\dagger}$} & \sigrob{$\text{4.29}^{\dagger}$} & \textcolor{gray}{3.20} & \textcolor{gray}{-4.84} & 5.93 & 9.50 \\
  \texttt{7} RobustViT \cite{chefer2022optimizing} & \sigrob{6.73} & \sigrob{1.13} & \textcolor{gray}{N/A} & \textcolor{gray}{N/A} & \textcolor{gray}{N/A} & \textcolor{gray}{N/A} \\
  \texttt{8} CLIP zero-shot \cite{radford2021learning, wortsman2022robust} & \sigrob{30.28} & \sigrob{10.79} & \textcolor{gray}{8.46} & \textcolor{gray}{-23.17} & \textcolor{gray}{-14.63} & \textcolor{gray}{-28.54} \\
\midrule
\textbf{Extreme Low-Shot} \\
\midrule
  \texttt{9} LP-FT \cite{kumar2022fine} & 3.71 & 1.75 & \textcolor{gray}{-0.62} & \textcolor{gray}{0.317} & 6.04 & 2.46 \\
  \texttt{10} \quad + CLIP & \sigrob{13.85} & \sigrob{4.51} & 3.59 & 6.24 & 9.30 & 8.35 \\
  \texttt{11} WiSE-FT \cite{wortsman2022robust} & 5.93 & 3.94 & \textcolor{gray}{-1.09} & \textcolor{gray}{0.00} & 5.62 & 2.44 \\
  \texttt{12} \quad + CLIP & \sigrob{29.90} & \sigrob{39.17} & \sigrob{6.87} & \sigrob{7.81} & \textcolor{gray}{-4.03} & \textcolor{gray}{-4.89} \\
  \texttt{13} Model Soups \cite{wortsman2022model} & 6.37 & 4.41 & \textcolor{gray}{-1.74} & \textcolor{gray}{-0.37} & 5.93 & 2.93 \\
  \texttt{14} \quad + CLIP & \sigrob{14.60} & \sigrob{5.10} & 0.56 & 2.63 & \sigrob{6.59} & \sigrob{9.64} \\
  \texttt{15} RobustViT \cite{chefer2022optimizing} & \sigrob{6.82} & \sigrob{5.32} & \textcolor{gray}{N/A} & \textcolor{gray}{N/A} & \textcolor{gray}{N/A} & \textcolor{gray}{N/A} \\
  \texttt{16} CLIP zero-shot \cite{radford2021learning, wortsman2022robust} &  \sigrob{30.28} & \sigrob{38.68} & 8.46 & 2.59 & \textcolor{gray}{-14.63} & \textcolor{gray}{-27.41} \\
  \midrule
\textbf{Moderate Low-Shot} \\
\midrule
  \texttt{17} LP-FT \cite{kumar2022fine} & 0.28 & 1.97 & \textcolor{gray}{-0.27} & \textcolor{gray}{2.62} & \textcolor{gray}{-0.01} & \textcolor{gray}{-3.15} \\
  \texttt{18} \quad + CLIP & \sigrob{17.76} & \sigrob{15.57} & \textcolor{gray}{-0.46} & \textcolor{gray}{3.82} & \textcolor{gray}{0.07} & \textcolor{gray}{-3.20} \\
  \texttt{19} WiSE-FT \cite{wortsman2022robust} & 3.25 & 4.90 & 3.51 & 3.96 & \textcolor{gray}{-0.37} & \textcolor{gray}{-2.77} \\
  \texttt{20} \quad + CLIP & \sigrob{29.22} & \sigrob{33.99} & \sigrob{7.81} & \sigrob{10.55} & \sigrob{7.61} & \sigrob{7.51} \\
  \texttt{21} Model Soups \cite{wortsman2022model} & 3.06 & 4.58 & 2.12 & 2.99 & \textcolor{gray}{-0.17} & \textcolor{gray}{-1.96} \\
  \texttt{22} \quad + CLIP & \sigrob{21.37} & \sigrob{17.82} & \textcolor{gray}{-0.24} & \textcolor{gray}{1.39} & \textcolor{gray}{4.22} & \textcolor{gray}{-0.77} \\
  \texttt{23} RobustViT \cite{chefer2022optimizing} & \sigrob{4.38} & \sigrob{5.70} & \textcolor{gray}{N/A} & \textcolor{gray}{N/A} & \textcolor{gray}{N/A} & \textcolor{gray}{N/A} \\
  \texttt{24} CLIP zero-shot \cite{radford2021learning, wortsman2022robust} &  \sigrob{30.28} & \sigrob{33.21} & \textcolor{gray}{8.46} & \textcolor{gray}{-4.45} & \textcolor{gray}{-14.63} & \textcolor{gray}{-27.41} \\
  \midrule
\textbf{High Low-Shot} \\
\midrule
  \texttt{25} LP-FT \cite{kumar2022fine} & \textcolor{gray}{-0.39} & \textcolor{gray}{2.70} & \textcolor{gray}{-0.98} & \textcolor{gray}{6.21} & 2.14 & 0.99 \\
  \texttt{26} \quad + CLIP & \sigrob{17.12} & \sigrob{19.11} & 1.62 & 6.38 & \textcolor{gray}{-2.39} & \textcolor{gray}{-5.53} \\
  \texttt{27} WiSE-FT \cite{wortsman2022robust} & 2.24 & 5.44 & \textcolor{gray}{-2.93} & \textcolor{gray}{3.65} & 2.34 & 1.87 \\
  \texttt{28} \quad + CLIP & \sigrob{28.20} & \sigrob{32.77} & \sigrob{4.35} & \sigrob{11.92} & \sigrob{6.81} & \sigrob{10.55} \\
  \texttt{29} Model Soups \cite{wortsman2022model} & 2.21 & 5.27 & \textcolor{gray}{-0.41} & \textcolor{gray}{5.57} & 2.72 & 2.84 \\
  \texttt{30} \quad + CLIP & \sigrob{21.65} & \sigrob{21.94} & 0.18 & 1.48 & 5.40 & 4.50 \\
  \texttt{31} RobustViT \cite{chefer2022optimizing} & \sigrob{2.68} & \sigrob{5.51} & \textcolor{gray}{N/A} & \textcolor{gray}{N/A} & \textcolor{gray}{N/A} & \textcolor{gray}{N/A} \\
  \texttt{32} CLIP zero-shot \cite{radford2021learning, wortsman2022robust} &  \sigrob{30.28} & \sigrob{31.79} & \textcolor{gray}{8.46} & \textcolor{gray}{-6.643} & \textcolor{gray}{-14.63} & \textcolor{gray}{-25.83} \\
\bottomrule \\
\end{tabular}
\caption{\small\textbf{Robustness intervention comparison.
}
The table shows effective ($\rho$) and relative ($\tau$) robustness of different interventions in the full-shot and low-shot regimes. $*$ and $\dagger$ denote numbers obtained from papers for ViTB-16 and ViTB-32 architecture respectively. Interventions that do not improve robustness in the full-shot regime are shown in \textcolor{gray}{gray}, while interventions that do so are shown in black. Interventions that significantly improve robustness in \textit{both} the full-shot regime and majority of low-shot regimes are highlighted in \sigrob{blue} for each dataset. 
Most interventions significantly improve robustness on ImageNet but not on other datasets. Only WiSE-FT with CLIP significantly improves robustness across datasets and data regimes. Absolute performances for computing $\tau$ are shown in table~\ref{tab:appx_rr_baselines}.
}
\label{tab:appx_ir_full_low_all}
\vspace{-10pt}
\end{table*}

\begin{figure*}[!htb]
    \centering
    \includegraphics[width=\linewidth]{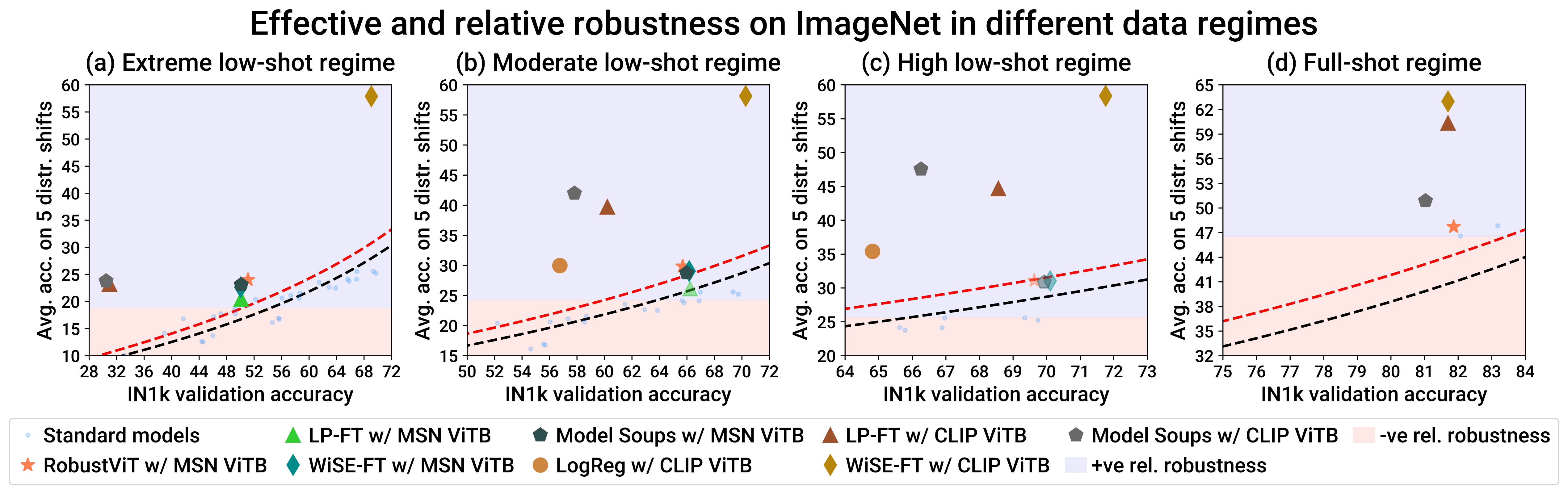}
    \caption{\small\textbf{Effect of robustness interventions on ImageNet.} Plots (a), (b), and (c) show performance of interventions in low-shot regimes (see table \ref{tab:appx_std_model_regimes}).
    Plot (d) shows performance of interventions in the full-shot regime.
    Interventions located above the black line ($\rho > 0$) and in the blue region ($\tau > 0$) are said to improve robustness. Interventions located above the red line and in the blue region are said to \textit{significantly} improve robustness (see Sec.~\ref{sec:appx_significance}).
    Interventions that significantly improve robustness are shown as opaque, whereas the ones that only improve robustness are shown as translucent.
    Most interventions significantly improve robustness across data regimes.
    }
    \label{fig:appx_ir_inet}
    \vspace{-7pt}
\end{figure*}

\begin{figure*}[!htb]
    \centering
    \includegraphics[width=\linewidth]{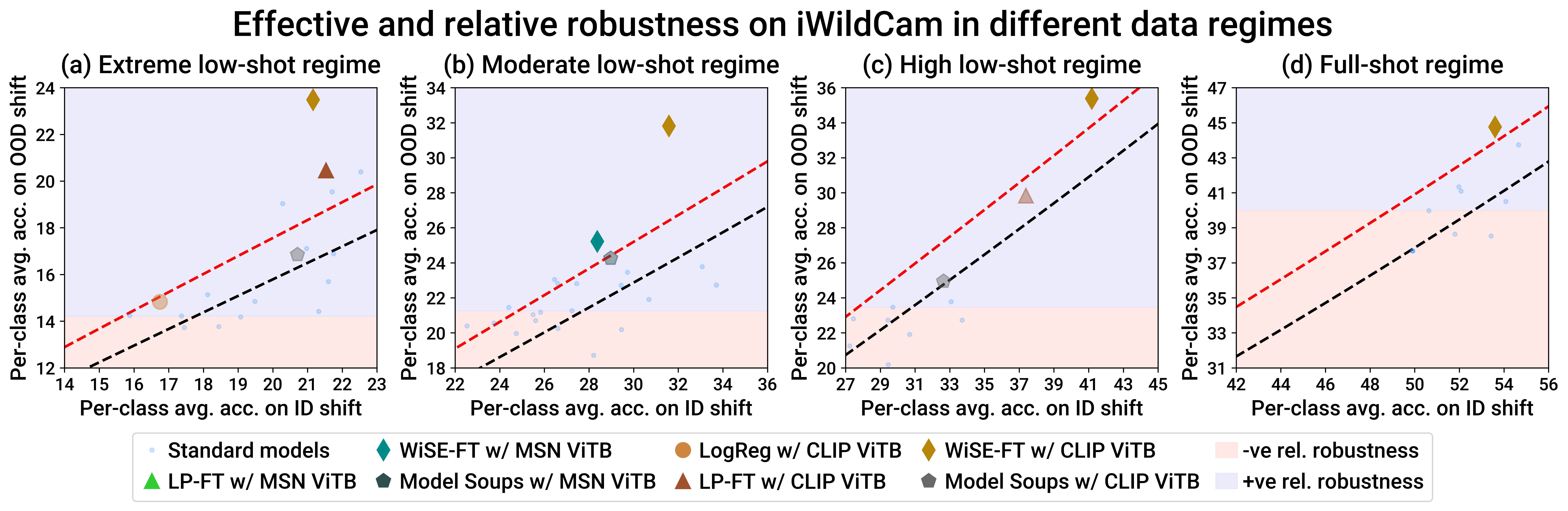}
    \caption{\small\textbf{Effect of robustness interventions on iWildCam.} Interventions often fail to improve robustness in both the full and low-shot regimes with MSN ViTB-16. Only WiSE-FT with CLIP ViTB-16 significantly improves robustness in all data regimes.
    }
    \label{fig:appx_ir_iwc}
    \vspace{-7pt}
\end{figure*}

\begin{figure*}[!htb]
    \centering
    \includegraphics[width=\linewidth]{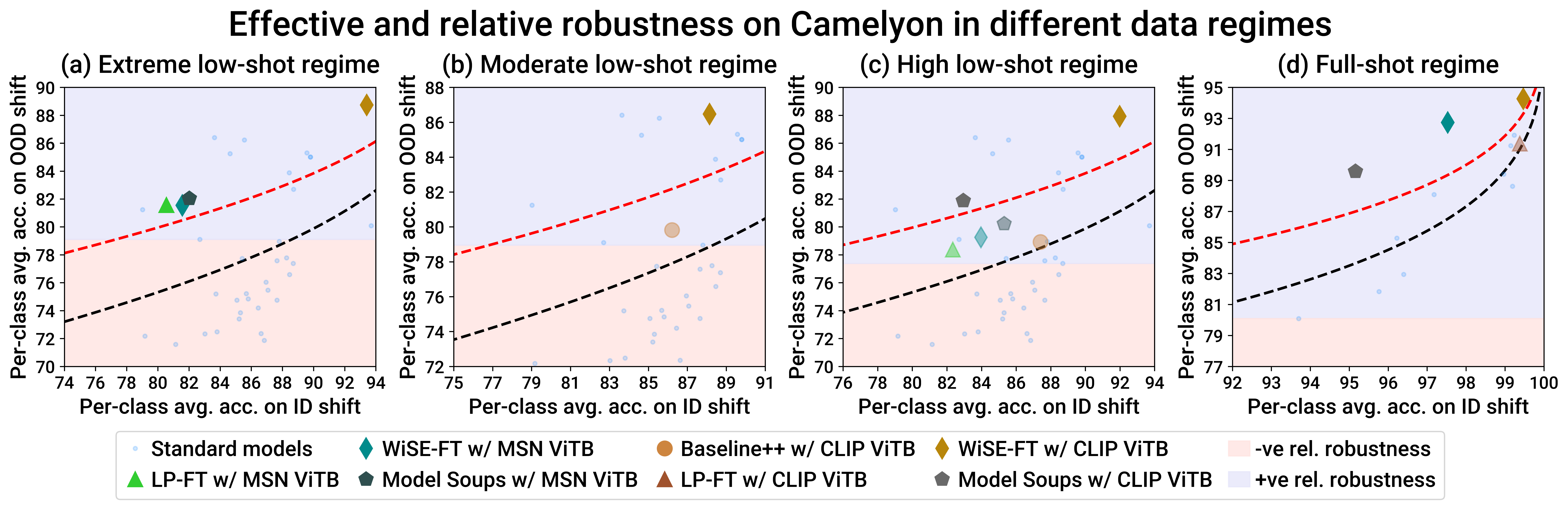}
    \caption{\small\textbf{Effect of robustness interventions on Camelyon.} Interventions often improve robustness in the full-shot regime with both MSN and CLIP ViTB-16 but fail to do so in \textit{extreme} or \textit{moderate} low-shot regimes for these models.
    Only WiSE-FT with CLIP significantly improves robustness across data regimes.
    Table \ref{tab:appx_ir_full_low_all} shows effective and relative robustness of interventions for further comparison.
    }
    \label{fig:appx_ir_cmlyn}
    \vspace{-7pt}
\end{figure*}

\begin{table*}[t]
\centering
\begin{tabular}{@{} l rr rr rr @{} }
\toprule
 & \multicolumn{2}{c}{\textbf{ImageNet}} & \multicolumn{2}{c}{\textbf{iWildCam}} & \multicolumn{2}{c}{\textbf{Camelyon}} \\ 
 \cmidrule(lr){2-3} \cmidrule(lr){4-5} \cmidrule(lr){6-7} 
 & \multicolumn{1}{c}{$\rho \uparrow$} & \multicolumn{1}{c}{$\tau \uparrow$} & \multicolumn{1}{c}{$\rho \uparrow$} & \multicolumn{1}{c}{$\tau \uparrow$} & \multicolumn{1}{c}{$\rho \uparrow$} & \multicolumn{1}{c}{$\tau \uparrow$} \\
\midrule
\textbf{Full-Shot Regime} \\
\midrule
  \texttt{1} LP-FT \cite{kumar2022fine} & \textcolor{gray}{6.83} & \textcolor{gray}{-0.57} & \textcolor{gray}{2.06} & \textcolor{gray}{-0.34} & 1.44 & 8.57 \\
  \texttt{2} \quad + CLIP & \sigrob{19.60$^*$} & \sigrob{12.52$^*$} & \textcolor{gray}{-3.60} & \textcolor{gray}{-9.09} & 0.37 & 9.54 \\
  \texttt{3} WiSE-FT \cite{wortsman2022robust} & 9.19 & -19.16 & \textcolor{gray}{1.85} & \textcolor{gray}{-5.06} & 4.08 & 10.13 \\
  \texttt{4} \quad + CLIP & \sigrob{22.24$^*$} & \sigrob{15.16$^*$} & \sigrob{3.98} & \sigrob{1.77} & \sigrob{2.85} & \sigrob{12.44} \\
  \texttt{5} Model Soups + CLIP \cite{wortsman2022robust} & \sigrob{11.00$^{\dagger}$} & \sigrob{3.04$^{\dagger}$} & \textcolor{gray}{3.20} & \textcolor{gray}{-7.84} & 5.93 & 7.75 \\
  \texttt{7} RobustViT \cite{chefer2022optimizing} &\sigrob{6.34} & \sigrob{0.87} & \textcolor{gray}{N/A} & \textcolor{gray}{N/A} & \textcolor{gray}{N/A} & \textcolor{gray}{N/A} \\
  \texttt{8} CLIP zero-shot \cite{radford2021learning, wortsman2022robust} & \sigrob{30.28} & \sigrob{10.79} & \textcolor{gray}{8.46} & \textcolor{gray}{-23.17} & \textcolor{gray}{-14.63} & \textcolor{gray}{-28.54} \\
\midrule
\textbf{Extreme Low-Shot} \\
\midrule
  \texttt{9} LP-FT \cite{kumar2022fine} & 7.10 & 2.95 & 2.04 & 4.59 & \textcolor{gray}{9.23} & \textcolor{gray}{-0.97} \\
  \texttt{10} \quad + CLIP & \sigrob{13.85} & \sigrob{6.37} & 3.56 & 6.69 & \textcolor{gray}{-4.03} & \textcolor{gray}{-12.20} \\
  \texttt{11} WiSE-FT \cite{wortsman2022robust} & 7.34 & 3.08 & 0.52 & 2.86 & \textcolor{gray}{9.66} & -\textcolor{gray}{0.71} \\
  \texttt{12} \quad + CLIP & \sigrob{29.90} & \sigrob{41.04} & \sigrob{6.87} & \sigrob{9.71} & \sigrob{6.59} & \sigrob{2.23} \\
  \texttt{13} Model Soups + CLIP \cite{wortsman2022model} & \sigrob{14.60} & \sigrob{6.97} & 0.56 & 3.08 & \textcolor{gray}{2.54} & \textcolor{gray}{-10.09} \\
  \texttt{15} RobustViT \cite{chefer2022optimizing} & \sigrob{8.95} & \sigrob{5.41} & \textcolor{gray}{N/A} & \textcolor{gray}{N/A} & \textcolor{gray}{N/A} & \textcolor{gray}{N/A} \\
  \texttt{16} CLIP zero-shot \cite{radford2021learning, wortsman2022robust} &  \sigrob{30.28} & \sigrob{38.68} & 8.46 & 2.59 & \textcolor{gray}{-14.63} & \textcolor{gray}{-27.41} \\
  \midrule
\textbf{Moderate Low-Shot} \\
\midrule
  \texttt{17} LP-FT \cite{kumar2022fine} & 5.45 & 5.14 & 0.49 & 5.22 & \textcolor{gray}{4.83} & \textcolor{gray}{-4.37} \\
  \texttt{18} \quad + CLIP & \sigrob{17.76} & \sigrob{16.16} & \textcolor{gray}{-0.46} & \textcolor{gray}{3.17} & \textcolor{gray}{0.07} & \textcolor{gray}{-8.12} \\
  \texttt{19} WiSE-FT \cite{wortsman2022robust} & 7.16 & 6.10 & \textcolor{gray}{-0.61} & \textcolor{gray}{4.50} & \textcolor{gray}{6.56} & \textcolor{gray}{-2.32} \\
  \texttt{20} \quad + CLIP & \sigrob{29.22} & \sigrob{34.58} & \sigrob{7.81} & \sigrob{9.90} & \sigrob{7.61} & \sigrob{2.59} \\
  \texttt{21} Model Soups + CLIP \cite{wortsman2022model} & \sigrob{21.37} & \sigrob{18.41} & \textcolor{gray}{-0.24} & \textcolor{gray}{0.74} & \textcolor{gray}{4.22} & \textcolor{gray}{-5.69} \\
  \texttt{23} RobustViT \cite{chefer2022optimizing} & \sigrob{8.39} & \sigrob{7.77} & \textcolor{gray}{N/A} & \textcolor{gray}{N/A} & \textcolor{gray}{N/A} & \textcolor{gray}{N/A} \\
  \texttt{24} CLIP zero-shot \cite{radford2021learning, wortsman2022robust} &  \sigrob{30.28} & \sigrob{33.21} & \textcolor{gray}{8.46} & \textcolor{gray}{-4.45} & \textcolor{gray}{-14.63} & \textcolor{gray}{-27.41} \\
  \midrule
\textbf{High Low-Shot} \\
\midrule
  \texttt{25} LP-FT \cite{kumar2022fine} & 3.61 & 4.71 & 1.51 & 6.44 & \textcolor{gray}{4.33} & \textcolor{gray}{-2.51} \\
  \texttt{26} \quad + CLIP & \sigrob{17.12} & \sigrob{19.15} & 1.62 & 6.06 & \textcolor{gray}{-2.39} & \textcolor{gray}{-10.84} \\
  \texttt{27} WiSE-FT \cite{wortsman2022robust} & 4.99 & 5.87 & 2.76 & 5.57 & \textcolor{gray}{4.66} & \textcolor{gray}{-2.25} \\
  \texttt{28} \quad + CLIP & \sigrob{28.20} & \sigrob{32.81} & \sigrob{4.35} & \sigrob{11.60} & \sigrob{6.81} & \sigrob{5.24} \\
  \texttt{29} Model Soups + CLIP \cite{wortsman2022model} & \sigrob{21.65} & \sigrob{21.98} & 0.18 & 1.16 & \textcolor{gray}{5.40} & \textcolor{gray}{-0.81} \\
  \texttt{31} RobustViT \cite{chefer2022optimizing} & \sigrob{6.93} & \sigrob{8.42} & \textcolor{gray}{N/A} & \textcolor{gray}{N/A} & \textcolor{gray}{N/A} & \textcolor{gray}{N/A} \\
  \texttt{32} CLIP zero-shot \cite{radford2021learning, wortsman2022robust} &  \sigrob{30.28} & \sigrob{31.79} & \textcolor{gray}{8.46} & \textcolor{gray}{-6.643} & \textcolor{gray}{-14.63} & \textcolor{gray}{-25.83} \\
\bottomrule \\
\end{tabular}
\caption{\small\textbf{Robustness intervention comparison with DINO ViTB \cite{caron2021emerging} as reference.}
The table shows effective ($\rho$) and relative ($\tau$) robustness of different interventions in the full-shot and low-shot regimes when applied on DINO ViTB-16. 
$*$ and $\dagger$ denote numbers obtained from papers for ViTB-16 and ViTB-32 architecture respectively. Interventions that do not improve robustness in the full-shot regime are shown in \textcolor{gray}{gray}, while interventions that do so are shown in black. Interventions that significantly improve robustness in \textit{both} the full-shot regime and majority of low-shot regimes are highlighted in \sigrob{blue} for each dataset. 
As with MSN (see table~\ref{tab:appx_ir_full_low_all}), most interventions significantly improve robustness on ImageNet but not on other datasets. 
Only WiSE-FT with CLIP significantly improves robustness across datasets and data regimes.
Absolute performances for computing $\tau$ are shown in table~\ref{tab:appx_rr_baselines}.
}
\label{tab:appx_ir_full_low_all_addon}
\vspace{40pt}
\end{table*}

\begin{figure*}[!htb]
    \centering
    \includegraphics[width=\linewidth]{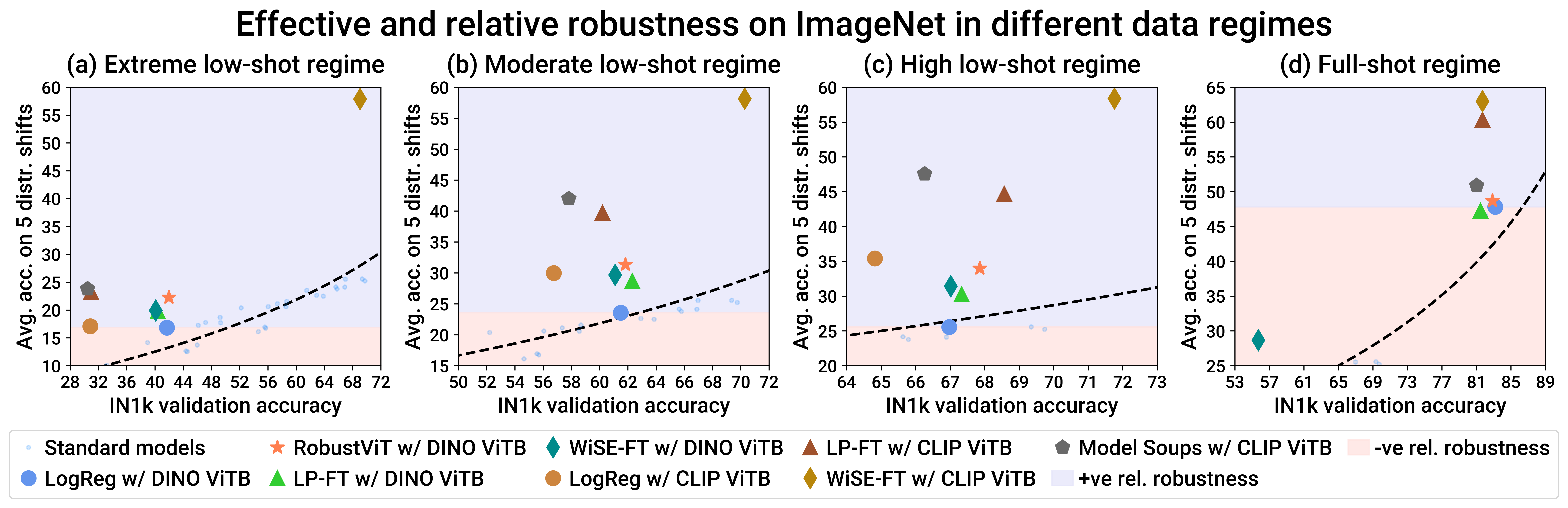}
    \caption{\small\textbf{Effect of robustness interventions on ImageNet with DINO \cite{caron2021emerging} as reference.} Plots (a), (b), and (c) show performance of interventions in low-shot regimes (see table \ref{tab:appx_std_model_regimes}).
    Plot (d) shows performance of interventions in the full-shot regime.
    Interventions located above the black line ($\rho > 0$) and in the blue region ($\tau > 0$) are said to improve robustness. 
    Interventions largely improve robustness in low-shot regimes with DINO ViTB-16 and in all data regimes when coupled with CLIP ViTB-16.
    }
    \label{fig:appx_ir_inet_addon}
\end{figure*}

\begin{figure*}[!htb]
    \centering
    \includegraphics[width=\linewidth]{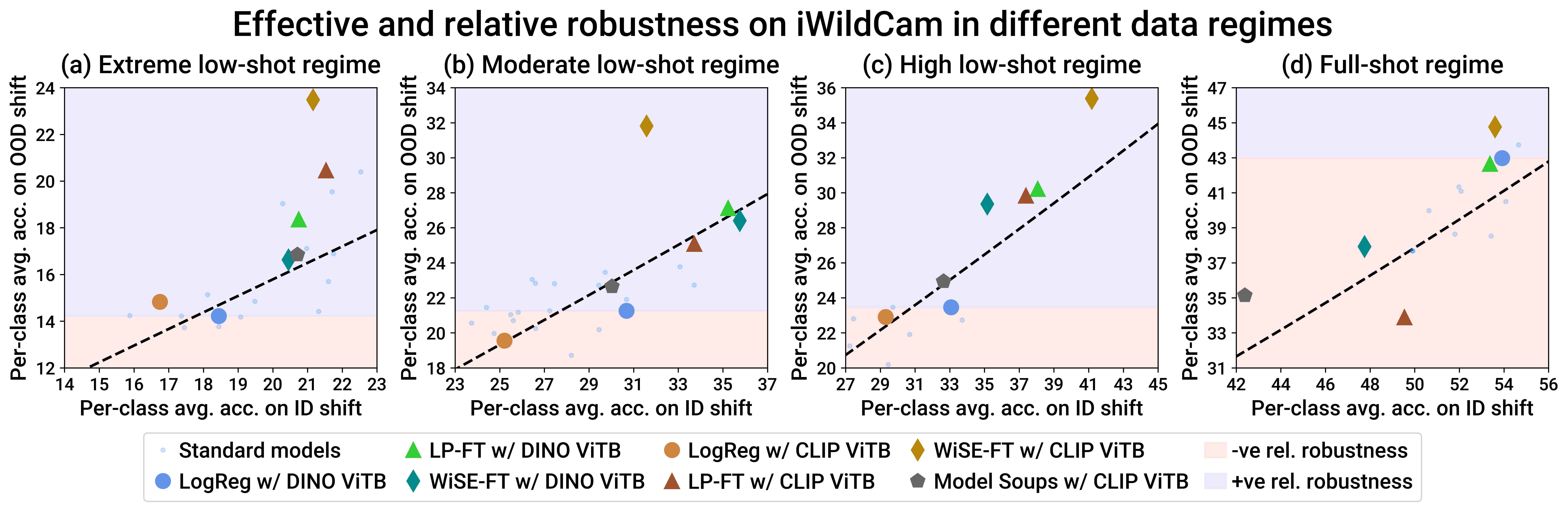}
    \caption{\small\textbf{Effect of robustness interventions on iWildCam with DINO \cite{caron2021emerging} as reference.} Interventions often improve robustness in the low-shot regimes but not in the full-shot regime with DINO. Only WiSE-FT with CLIP improves robustness in all data regimes.
    }
    \label{fig:appx_ir_iwc_addon}
\end{figure*}

\begin{figure*}[!htb]
    \centering
    \includegraphics[width=\linewidth]{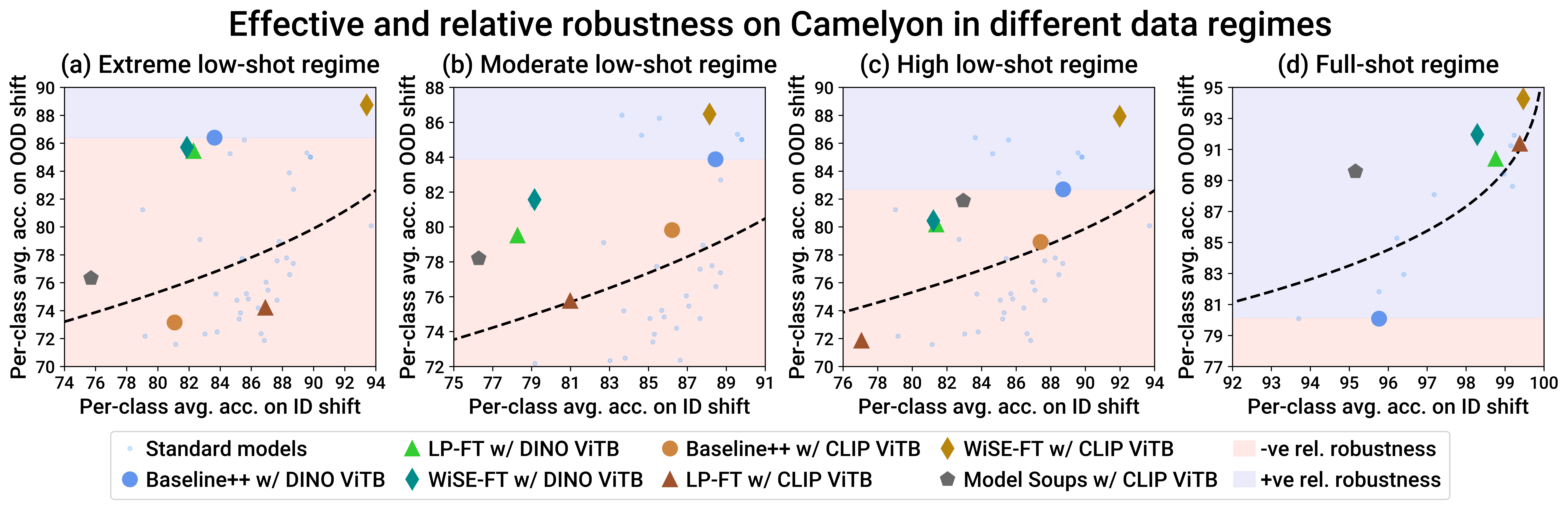}
    \caption{\small\textbf{Effect of robustness interventions on Camelyon with DINO \cite{caron2021emerging} as reference.} Interventions often improve robustness in the full-shot regime with both DINO and CLIP ViTB-16 but often fail to do so in the low-shot regimes,
    except WiSE-FT with CLIP.
    Table \ref{tab:appx_ir_full_low_all_addon} shows effective and relative robustness of interventions with DINO ViTB-16 for further comparison.
    }
    \label{fig:appx_ir_cmlyn_addon}
\end{figure*}

We show the effective and relative robustness of the interventions in all datasets and data regimes in table~\ref{tab:appx_ir_full_low_all}. By default, we use MSN \cite{assran2022masked} as reference and ViTB-16 models for applying interventions. To complement these results and our findings in the main paper, we obtain the curve $\beta_\lambda(x)$ (see Eq.~\ref{eq:appx_beta_lambda}) for measuring significance. Table~\ref{tab:appx_beta_lambda_params} shows the obtained parameter values for the different datasets.

We summarize the results for ImageNet in Fig.~\ref{fig:appx_ir_inet}, iWildCam in Fig.~\ref{fig:appx_ir_iwc}, and Camelyon in Fig.~\ref{fig:appx_ir_cmlyn}. While most interventions significantly improve robustness on ImageNet across data regimes, they fail to do so on iWildCam and Camelyon datasets.
WiSE-FT with CLIP is the only intervention which significantly improves robustness across the different datasets and data regimes.

For completeness, we also report the mean and standard deviation of some interventions with CLIP across $2$ different runs on iWildCam and Camelyon datasets in table~\ref{tab:appx_perf_var}. It can be seen that OOD variation can be high even when ID variation is small, as also observed by \cite{wortsman2022robust}. Surprisingly, Model Soups generally exhibits the smallest variance even though it's hyperparameters are sampled randomly as shown in table~\ref{tab:appx_modelsoups_hparams}. However, WiSE-FT often leads to much better performance with relatively small variance.

\begin{table}[t]
\centering
\begin{tabular}{@{}lccc@{}}
\toprule
\multirow{2}{*}{\textbf{Dataset}} & \multicolumn{3}{c}{\textbf{Parameters for \boldmath$\beta_{\lambda}(x)$}} \\ \cmidrule(l){2-4}
 & $w$ & $b$ & $d$ \\ \midrule
ImageNet~\cite{deng2009imagenet} & 0.825 & -1.609 & 0.136 \\
iWildCam~\cite{beery2020iwildcam} & 0.850 & -0.496 & 0.128 \\
Camelyon~\cite{bandi2018detection} & 0.325 & 0.665 & 0.268 \\ \bottomrule
\end{tabular}
\caption{\small\textbf{Parameters for \boldmath$\beta_{\lambda}(x)$.} For each dataset, we list the values for $w$, $b$, and $d$ to obtain the function $\beta_{\lambda}(x)$ (see Eq.~\ref{eq:appx_beta_lambda})}
\label{tab:appx_beta_lambda_params}
\end{table}

\begin{table}[t]
\centering
\resizebox{\columnwidth}{!}{
\begin{tabular}{@{}lllllll@{}}
\toprule
\multirow{2}{*}{\textbf{Data Regime}} & \multicolumn{2}{c}{\textbf{ImageNet}} & \multicolumn{2}{c}{\textbf{iWildCam}} & \multicolumn{2}{c}{\textbf{Camelyon}} \\ 
 \cmidrule(lr){2-3} \cmidrule(lr){4-5} \cmidrule(lr){6-7}
                                      & \multicolumn{1}{c}{MSN}    & DINO     & MSN      & \multicolumn{1}{c}{DINO}   & MSN               & DINO              \\ \midrule
Full-Shot Regime                      & 46.57                      & 47.82    & 39.98    & 42.99                      & 80.09             & 81.83             \\
Extreme Low-Shot                      & 18.69                      & 14.15    & 14.22    & 13.77                      & 79.10             & 86.41             \\
Moderate Low-Shot                     & 24.16                      & 20.60    & 21.26    & 21.91                      & 78.96             & 83.88             \\
High Low-Shot                         & 25.58                      & 22.51    & 23.46    & 23.78                      & 77.38             & 82.69             \\ \bottomrule
\end{tabular}}
\caption{\small\textbf{
OOD performances of reference models.
} 
The table shows the OOD performances of MSN and DINO ViTB-16 used to compute relative robustness $\tau$ in tables \ref{tab:appx_ir_full_low_all} and \ref{tab:appx_ir_full_low_all_addon}.
}
\label{tab:appx_rr_baselines}
\end{table}

\begin{table}[t]
\centering
\resizebox{\columnwidth}{!}{
\begin{tabular}{@{}lllll@{}}
\toprule
\multirow{2}{*}{\textbf{Data Regime}} &
\multicolumn{2}{c}{\textbf{iWildCam}} & \multicolumn{2}{c}{\textbf{Camelyon}} \\
\cmidrule(lr){2-3} \cmidrule(lr){4-5}
& ID & OOD & ID & OOD \\
\midrule
\textbf{Full-Shot} \\
\midrule
WiSE-FT + CLIP & 53.18 $\pm$ 0.42 & 44.92 $\pm$ 0.16 & 99.46 $\pm$ 0.01 & 94.41 $\pm$ 0.14 \\
LP-FT + CLIP & 49.85 $\pm$ 0.31 & 33.89 $\pm$ 1.78 & 99.22 $\pm$ 0.16 & 87.71 $\pm$ 3.65 \\
Model Soups + CLIP & 42.39 $\pm$ 0.00 & 35.14 $\pm$ 0.00 & 95.17 $\pm$ 0.01 & 89.58 $\pm$ 0.01 \\
\midrule
\textbf{Extreme Low-Shot} \\
\midrule
WiSE-FT + CLIP & 19.81 $\pm$ 1.36 & 22.89 $\pm$ 0.59 & 93.17 $\pm$ 0.24 & 88.91 $\pm$ 0.17 \\
LP-FT + CLIP & 19.86 $\pm$ 1.68 & 19.88 $\pm$ 0.58 & 87.57 $\pm$ 0.66 & 80.80 $\pm$ 6.59 \\
Model Soups + CLIP & 20.70 $\pm$ 0.01 & 16.84 $\pm$ 0.01 & 75.73 $\pm$ 0.01 & 76.18 $\pm$ 0.10 \\
\midrule
\textbf{Moderate Low-Shot} \\
\midrule
WiSE-FT + CLIP & 31.75 $\pm$ 0.16 & 31.57 $\pm$ 0.25 & 89.25 $\pm$ 1.11 & 86.83 $\pm$ 0.36 \\
LP-FT + CLIP & 32.64 $\pm$ 1.09 & 23.93 $\pm$ 1.16 & 81.27 $\pm$ 0.29 & 76.78 $\pm$ 1.02 \\
Model Soups + CLIP & 28.28 $\pm$ 1.27 & 22.65 $\pm$ 0.31 & 76.65 $\pm$ 0.26 & 78.71 $\pm$ 0.36 \\
\midrule
\textbf{High Low-Shot} \\
\midrule
WiSE-FT + CLIP & 41.70 $\pm$ 0.52 & 35.44 $\pm$ 0.06 & 91.03 $\pm$ 0.95 & 87.78 $\pm$ 0.16 \\
LP-FT + CLIP & 37.09 $\pm$ 0.3 & 29.78 $\pm$ 0.07 & 76.13 $\pm$ 0.94 & 71.03 $\pm$ 0.82 \\
Model Soups + CLIP & 31.29 $\pm$ 0.98 & 25.09 $\pm$ 0.11 & 80.95 $\pm$ 1.91 & 81.03 $\pm$ 0.60 \\
\bottomrule
\end{tabular}}
\caption{\small\textbf{Performance variance.}~We report the mean and std. deviation of some interventions with CLIP across $2$ runs. Model Soups generally exhibits the smallest variance but WiSE-FT often leads to much better performance with relatively small variance.
}
\label{tab:appx_perf_var}
\end{table}

\section{Results for other initializations.}
One might ask how dependent our observations are on the choice of the reference model, i.e. MSN ViTB-16 and whether other initializations result in the same set of observations. To answer this, we apply the interventions described in Sec.~\ref{sec:appx_interventions} on DINO ViTB-16. The absolute out-of-distribution (OOD) performances with both models are shown in table~\ref{tab:appx_rr_baselines}. We omit Model Soups with DINO from this experiment due to limited compute. The dataset-wise observations are described below.

\textbf{ImageNet.} We show the results of this experiment in Fig.~\ref{fig:appx_ir_inet_addon}. Similar to the findings for MSN, interventions are largely effectively and relatively robust in the low-shot regimes when coupled with DINO. RobustViT also improves robustness in all data regimes. With CLIP ViTB-16, intervention are effectively and relatively robust in all data regimes. As shown in table~\ref{tab:appx_ir_full_low_all_addon}, zero-shot CLIP improves robustness on ImageNet but often fails to do so on other datasets and data regimes.

\textbf{iWildCam.} We show the results of this experiment in Fig.~\ref{fig:appx_ir_iwc_addon}. With DINO, interventions are often effectively and relatively robust in the low-shot regimes but neither effectively nor relatively robust in the full-shot regime. As with MSN, WiSE-FT with CLIP is the only intervention which improves robustness in all data regimes.

\textbf{Camelyon.} We show the results of this experiment in Fig.~\ref{fig:appx_ir_cmlyn_addon}. As with MSN, most interventions improve robustness in the full-shot regime and WiSE-FT with CLIP does so in all data regimes. However, unlike MSN, other interventions fail to be relatively robust in all low-shot regimes instead of just the \textit{extreme} or \textit{moderate} low-shot regimes.

To complement our findings, we also show the effective and relative robustness of the interventions on different datasets and data regimes in table~\ref{tab:appx_ir_full_low_all_addon}. We follow the same procedure for measuring significance as described in Sec.~\ref{sec:appx_significance}. Consistent with the findings for MSN, we see that (1) most interventions significantly improve robustness on ImageNet but not on other datasets and (2) no intervention significantly improves robustness across datasets and data regimes, except WiSE-FT with CLIP. Overall, our findings hold for multiple initializations and show that robustness to natural shifts on ImageNet and in full-shot regimes might not imply that on other datasets and in the low-shot regimes. 

\section{Related works}
\label{sec:appx_rel_works}
We describe additional related works that we were unable to include in the main paper due to space constraints.

\par\noindent
\textbf{Domain generalization.} 
In domain generalization, the goal is to generalize to an inaccessible target domain while assuming access to one or more fully labelled source domains \cite{blanchard2011generalizing, li18adversarial, gulrajani2020search, qiao2020learning, zhou21mixstyle, cha21swad, zhou2022domain}.
While recent methods often use vision-language models such as CLIP \cite{radford2021learning} for impressive robustness gains through strategic fine-tuning \cite{kumar2022fine} or weight-space ensembles \cite{wortsman2022robust, wortsman2022model}, they also rely on abundant labelled data for training which can be prohibitive for practitioners. Thus, we investigate the effectiveness of these methods in low-shot regimes on diverse datasets.

\par\noindent
\textbf{Domain adaptation.}
Domain adaptation (DA) seeks to transfer a model trained on a source domain to an unseen target domain.
When the target domain doesn't have labels, the setting is referred to as unsupervised domain adaptation (UDA) which has been extensively studied \cite{tzeng2014deep, long2015learning, ganin2015unsupervised, tzeng2017adversarial, long2018conditional, hoffman2018cycada, kim2022broad, yang2023tvt}. While a large body of works rely on supervised ImageNet initializations for UDA, some works have focused on self-supervised adaptation with CNNs \cite{kim2021cds, shen2022connect} and ViTs \cite{prabhu2022adapting}. Recent works have also studied test-time adaptation \cite{wang2020tent, liu2021ttt++, wang2022continual} which focuses on online learning, and few-shot adaptation \cite{motiian2017few, zhao2021domain, zhang2022few, yazdanpanah2022visual} which is often similar to the CD-FSL setting. Crucially, robustness studies and our study differs from DA and these works by \textit{not} assuming access to the target data.

\end{document}